\documentclass[lettersize,journal,twoside]{IEEEtran}
\usepackage{amsmath,amsfonts}
\usepackage{algorithmic}
\usepackage{algorithm}
\usepackage{array}
\usepackage{tabularx}
\usepackage[caption=false,font=normalsize,labelfont=sf,textfont=sf]{subfig}
\usepackage{textcomp}
\usepackage{stfloats}
\usepackage{url}
\usepackage{verbatim}
\usepackage{graphicx}
\usepackage{cite}
\usepackage{multirow}
\usepackage{caption}
\usepackage{subfig}
\usepackage{rotating}

\begin{document}

\title{Exploratory Driving Performance and Car-Following Modeling for Autonomous Shuttles Based on Field Data}

\author{Renan Favero,\IEEEmembership{ Student Member, IEEE
}, Lily Elefteriadou,\IEEEmembership{ Member, IEEE} 
\thanks{This paper was supproted by the U.S. DOT/Region 4
STRIDE University Transportation Center. Project H6 is
"STRIDE Project H-6".}
\thanks{The authors are with Department of Civil and Coastal Engineering, University of Florida, FL 32608 USA
 (e-mail: renanfavero@ufl.edu; elefter@ce.ufl.edu).}}


\markboth This work has been submitted to the IEEE for possible publication. Copyright may be transferred without notice, after which this version may no longer be accessible.


\maketitle

\begin{abstract}
Autonomous shuttles (AS) operate in several cities and have shown potential to improve the public transport network. However, there is no car-following model that is based on field data and allows decision-makers (planners, and traffic engineers) to assess and plan for AS operations. To fill this gap, this study collected field data from AS, analyzed their driving performance, and suggested changes in the AS trajectory model to improve passenger comfort. 
A sample was collected with more than 4,000 seconds of AS following a conventional car (human driver). The sample contained GPS positions from both AS and conventional vehicles.  Latitude and longitude positions were used to calculate the speed, acceleration, and jerk of the leader and follower. The data analyses indicated that AV has higher jerk values that may impact the passengers' comfort.  Several existing models were evaluated, and the researchers concluded that the calibrated ACC model resulted in lower errors for AS spacing and speed. The results of the calibration indicate that the AS has lower peak acceleration and higher deceleration than the parameters that were calibrated for autonomous vehicle models in other research.

\end{abstract}

\begin{IEEEkeywords}
Autonomous Shuttle, car-following, autonomous vehicle trajectories, IDM, ACC, IIDM.
\end{IEEEkeywords}

\section{Introduction}

\IEEEPARstart{A}{utonomous} shuttles (AS) are small self-driving vehicles and are classified as autonomous vehicles (AV) level-4 \cite{sae2018}. AS use on-board sensors and cameras, combine data, and processes it with machine learning algorithms to navigate roads, interact with vehicles, and avoid obstacles. AS have been deployed in several countries (USA, France, Switzerland, Germany, Austria, Norway, Canada, Australia, Belgium, and China \cite{Iclodean2020AutonomousReview, Feys2020ExperienceRegion, Piatkowski2021AutonomousThem}).

AS drive in predetermined routes mapped by GPS and travel according to input data received by sensors. Currently, AS travel at low speeds (max speeds are 17.6 - 22.0 $ft/s$) and are typically used for short trips within controlled environments (campuses, airports, or parks) \cite{Haque2020}. 

Most of the studies on AV suggest that they may affect traffic operations, including road capacity \cite{Calvert2017WillFlow},  road safety \cite{Papadoulis2019EvaluatingMotorways},  and urban sustainability \cite{Oikonomou2020}. Papadoulis \textit{et al.}\cite{Papadoulis2019EvaluatingMotorways} suggest that even for low market penetration, AV can improve road safety. Calvert \textit{et al.}\cite{Calvert2017WillFlow} indicate that for lower market penetration, AV can reduce road capacity, but with penetration rates greater than 70\%, they can improve traffic flow and increase capacity. AV also can improve the energy consumption and gas emission \cite{Lu2019Energy-EfficientVehicles, Liu2019CanEvaluation, Alkim2007FieldDriver}.

AS have vehicle specifications different from those of AV. They differ in passenger capacity (usually 8 to 15 passengers), vehicle performance (maximum speed, speed profile, acceleration), physical characteristics (weight, size, engine power), and in that their route is predetermined (including loading and unloading stops). Preliminary results indicate that AS might improve the mobility of people and goods \cite{Ainsalu2018StateBuses, Azad2019FullyDirections, Li2020DataCities} and can reduce CO2 emissions \cite{Oikonomou2020}.  We were not able to find any research exploring AS movement using field data, nor any research assessing their potential impact on the highway network in which they operate. Given the number of existing and planned AS deployments, it is important to be able to model their operational impact based on their performance.

 Car-following is the process in which the follower vehicle interacts with the leader vehicle. In this context, a well-calibrated car-following model for AS is essential for simulations, so that agencies can evaluate AS's expected impact under various scenarios. Moreover, defining a representative model to predict and assess traffic flow is essential, as it directly affects vehicle trajectories, highway capacity, and facility speed \cite{Elefteriadou2014ModelingVehicles}. 

To address this gap, the objectives of this paper are: (1) to obtain field data from an existing AS deployment, (2) to evaluate the extracted trajectories and AS operations under car-following conditions, (3) to select the most promising car-following models identified from the literature and calibrate them using the field data obtained, and (4) to compare the calibrated parameters of the AS to those of conventional vehicles and AV in the literature. 

To address the objectives of this study, this paper is organized into the following sections:  (III) Method; (IV) Results; and (V) Conclusions.

 \section{LITERATURE REVIEW}

The literature review describes (1) existing field deployments of AV and their findings, (2) describes exploratory analyses presented in previous studies about AS, (3) presents the models used to simulate AV, and (4) compares qualitatively the car-following models described in the literature. 
\\
\\
\subsection{Existing AV field deployments and their findings}

Studies investigating AS with field data are rather limited \cite{Shi2021PreliminaryShuttle}. Based on field AS trajectories, some studies presented preliminary evaluations. A recent study tested some components of an AS  \cite{Shi2021PreliminaryShuttle}. The results suggest that AS are safe for travel leading or following other vehicles, and can detect obstacles and pedestrians with adequate accuracy. In addition, significant differences were observed between the speed, acceleration, and stopping distance from trajectories carried out by a human driver compared to an AS.

A systematic review of the literature on AS pilot projects suggests the prospects for future AS pilot programs and their integration into existing transit systems \cite{MahmoodiNesheli2021DriverlessDeployment}. These pilots aim to operate under various road conditions, higher speeds, and more trafficked areas. Although public feedback has generally been positive, there is still skepticism about using shuttles without onboard operators \cite{MahmoodiNesheli2021DriverlessDeployment}.

Regarding AV, users driving with Adaptive Cruise Control (ACC) showed a significant effect on user driving behavior. The majority of drivers choose the shortest headway setting. Most of the participants stated that they are more likely to participate in secondary tasks while the ACC mode is on \cite{Alkim2007FieldDriver}. Field experiments on the longitudinal characteristics of human driver behavior when following an AV revealed that different drivers have different driving behaviors \cite{Zhao2020FieldVehicle}. The findings suggest that, depending on the characteristics of the drivers, classic car-following models need to be revised to consider traffic mixed with AV.

\subsection{AV movements through models and simulation}

In general, AV (passenger vehicles, not AV shuttles) have been simulated in microsimulation using two approaches \cite{Ahmed2019EvaluationVehicles, Papadoulis2019EvaluatingMotorways}. One approach is based on adjusting car-following parameters of models created to predict trajectories of human drivers, for example, the Wiedemann model. In this case, the default models are adapted to represent the deterministic behavior of AV and communication between vehicles
(V2V) \cite{DfT2016ResearchFlow, Morando2018StudyingMeasures, Stanek2018MeasuringSimulation, Sanusi2019MeasuringZones, SenturkBerktas2020EffectIntersections}. Another approach is based on developing a new logic for driving behavior and communication V2V. \cite{Shladover2012ImpactsFlow, Hamilton2018DedicatingVehicles, Shi2019CapacityMethod}.
 
Based on the first approach, the CoExist project used real data from AV and proposed a calibrated Wiedemann 99 model to simulate AV \cite{Sukennik2018Micro-simulationD2.5}. The use of a psychophysical model of AV has some limitations since it may not accurately represent an AV trajectory because it is based on assumptions related to human drivers' perception thresholds. This depends on the relative speed between vehicles; when it is small, the human driver is not able to perceive it and does not change the acceleration \cite{Hoogendoorn2014AutomatedReview}. Adapting stochastic models to AV's simulation can also be considered a limitation, as AV has a deterministic behavior; the simulation can result in traffic instability and unrealistic properties in the deterministic limit \cite{Treiber2013MicroscopicApproach}.
 
Based on the second approach, the researchers proposed external driver models to simulate a new logic for  AV trajectories in different software, such as VISSIM \cite{Li2019ParsimoniousTraffic, Papadoulis2019EvaluatingMotorways}, AIMSUN \cite{Roncoli2015ModelVehicles} and Transmodel \cite{Bao2020}. However, these models were not designed for AS, but only for AV passenger cars.
 
  In another project, the researchers used an in-built model with theoretic car-following parameters to investigate the AS operation \cite{Gasper2018SimulationSpace}. \cite{Oikonomou2020} evaluated the impacts of the AS on traffic, safety, and the environment using an ACC model. However, these studies did not use calibrated car-following models based on actual AS trajectories, and the parameters used were adopted based on the expected AV movement.

 With a different approach, the car-following control procedures were represented through the theoretical functioning of the sensor \cite{He2021DesignBuses, Li2021DevelopmentEnvironment, Weissensteiner2021VirtualShuttles}. Although the trajectories were calibrated with real data in these studies, the simulations did not take into account the interaction with other road users. In addition to that, studies focused on sensors' perception have a high computational cost. Then, software based on sensors' perception algorithms is usually not applied to assess the interaction of several vehicles simultaneously. Again, these studies did not consider larger AV shuttles. 
 
 In a previous study, the Intelligent Driver Model (IDM) model was calibrated for AS trajectories. However, this research project considered a small sample size and a specific scenario. Therefore, this project was unable to obtain the car-following model or identify a statistical difference between human driver behavior and AS trajectory due to a lack of  available data \cite{Maehara2021CALIBRATIONSHUTTLE}.
 
\subsection{Car-Following Models for Autonomous Vehicles (AV)}

This section briefly summarizes the most recent extensive literature on car-following. The IDM  is a deterministic model that describes vehicle acceleration as a function of acceptance, speed, and the speed difference between the leader and the follower \cite{Treiber2000CongestedSimulations}. Previous research indicated that IDM results in realistic behavior, and with a single model we can simulate different capacity flow for various networks \cite{Treiber2000CongestedSimulations}. This model is suitable to simulate AV since it takes into account human stochastic behavior. In addition, this model considers the speed and position of the leader vehicle to determine its acceleration. 

According to Treiber and Kesting \cite{Treiber2000CongestedSimulations} the IDM has the following characteristics: it is accident-free due to the dependence on relative speed, hysteresis effects can simulate complex states, all parameters have a reasonable interpretation and are empirically measurable, it was calibrated to empirical data, and the macroscopic version of the model is provided in \cite{Treiber2000CongestedSimulations}.
According to Kesting and Treiber \cite{Kesting2008CalibratingData}, in most situations, the IDM models simulate
acceleration and deceleration in a satisfactory manner. However, according to Kesting \textit{et. al} \cite{Kesting2010EnhancedCapacity} the IDM model can result in unrealistic deceleration when the vehicle gap is significantly lower than the desired gap \cite{Kesting2010EnhancedCapacity}. 

To overcome these limitations, the IDM model \cite{Kesting2010EnhancedCapacity} was improved considering two regimes: when the gap is greater than the desired gap (IIDM model), and also combined with constant acceleration heuristics (CAH) to limit unrealistic deceleration.
 The CAH model includes the following assumptions: (1) the leader will not change its acceleration suddenly in the following few seconds; (2) no safe time progress and minimum distance are required; and (3) the drivers react without delay (zero reaction time) \cite{Kesting2010EnhancedCapacity}.

The adaptive cruise control (ACC) car-following model was proposed, mainly to simulate vehicles initially with a lower level of autonomy\cite{Kesting2010EnhancedCapacity}. This model was suggested to eliminate unrealistic behaviors when vehicles follow with small gaps. 

A simplified version of the NISSAN ACC car was simulated \cite{Bu2010DesignSystem} to evaluate the increase in roadway capacity \cite{Shladover2012ImpactsFlow}. This ACC model includes fewer parameters than the IDM model, and the simulation results closely align with the experimental results of road tests conducted on production vehicles \cite{Milanes2014CooperativeSituations}.  However, the data collected for this model calibration in previous studies included limited and controlled vehicle interactions, which differs from real traffic. 
The cooperative ACC (CACC) model \cite{Milanes2014ModelingData} and MIXIC \cite{Melson2018DynamicControl} were proposed to simulate the cooperation capacity and information exchange between AVs \cite{Vanderwerf2002EffectsCapacity}. It was concluded that CACC can increase road capacity and reduce highway congestion, but naive deployment of CACC-exclusive lanes could cause an increase in total system travel time.

\subsubsection{Car-following Model comparison}
 
The IDM \cite{Treiber2000CongestedSimulations} and Velocity Difference Model (VDIFF) \cite{Jiang2001FullTheory} were compared using different error measures \cite{Kesting2008CalibratingData}. These car-following models have the same number of parameters, and the results showed that the IDM was more robust for reasonable parameter changes and resulted in significantly fewer prediction errors. However, VDIFF led to larger errors, and the calibrated parameter values of VDIFF strongly depend on the optimization criteria. Also, the VDIFF was more sensitive to parameter changes, and, in some cases, the model predicted collisions, which reflected a significant error \cite{Kesting2008CalibratingData}.
 
Using experimental data from AV (level 2 and 3) trajectories, researchers compared the IDM, ACC and CACC controller models \cite{Milanes2014ModelingData}. The results showed that IDM produces smoother behavior but with considerable variation in the clearance gap and slower response. The results obtained in this research project suggested that a new model for CACC and ACC was more accurate for AV simulation.

Based on human driver trajectories, an improved IDM (IIDM) , the IDM, and the MIXIC model were compared. According to the analysis, IIDM had the best performance. In IDM and MIXIC, the models presented higher calibrated parameters, and thus the models may result in extreme cases \cite{Kim2023IdentifyingHighways}.
 
Although there are a large number of studies related to AV operation \cite{Sukennik2018Micro-simulationD2.5, Tiernan2019CARMAApplications, Zhao2020FieldVehicle, Fremont2020FormalWorld}, analyses based on field data for full AVs (levels 4 - 5) are still limited. The main reason is the lack of real AV trajectories (levels 4 - 5) extracted in mixed traffic. 

This review of the literature examined a variety of models and identified the most suitable ones for calibration. It was concluded that the IDM, IIDM, and the ACC model proposed by \cite{Milanes2014ModelingData} are potentially suitable for calibrating AS car-following models. According to the literature, the results of these model calibrations present reasonable accuracy for predicted trajectories, include few variables, and are relatively flexible. 
The CACC model and MIXIC models were not calibrated because they simulate vehicles with communication, and the AS studied was not equipped with communication vehicle-to-infrastructure (V2I) or vehicle-to-vehicle (V2V) capabilities.

\section{METHOD}

This section describes the method for this study and is divided into the following subsections: data collection, data cleaning, exploratory analysis, description of car-following calibration, and comparison of calibrated parameters to those provided in the literature.

\subsection{Data collection}

\begin{figure}[!ht]
  \centering
  \includegraphics[width=0.5\textwidth]{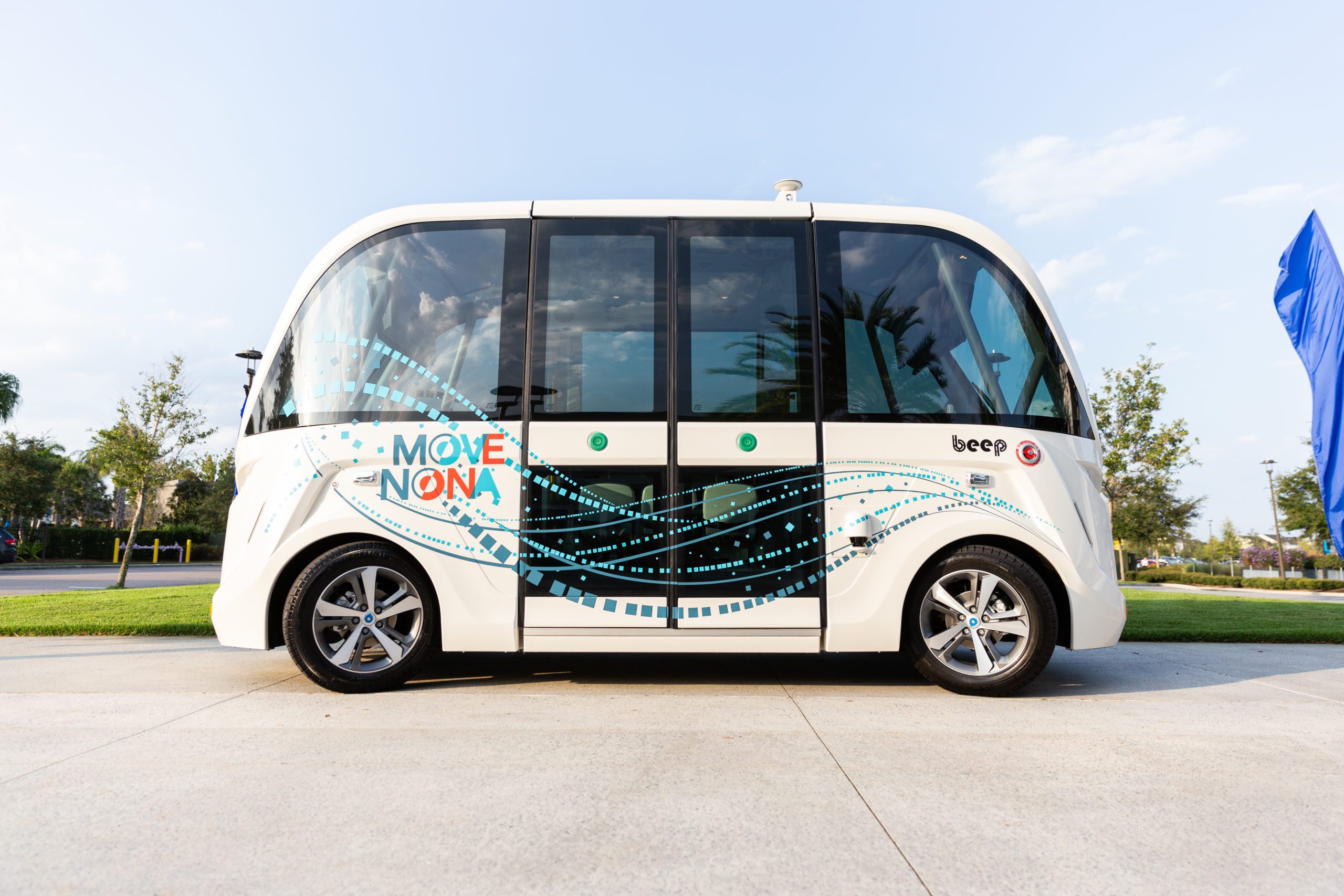}
  \caption{The first autonomous vehicles in Central Florida operating in Lake Nona}\label{fig:nona_av}
\end{figure}

\begin{figure}[!ht]
  \centering
  \includegraphics[width=0.5\textwidth]{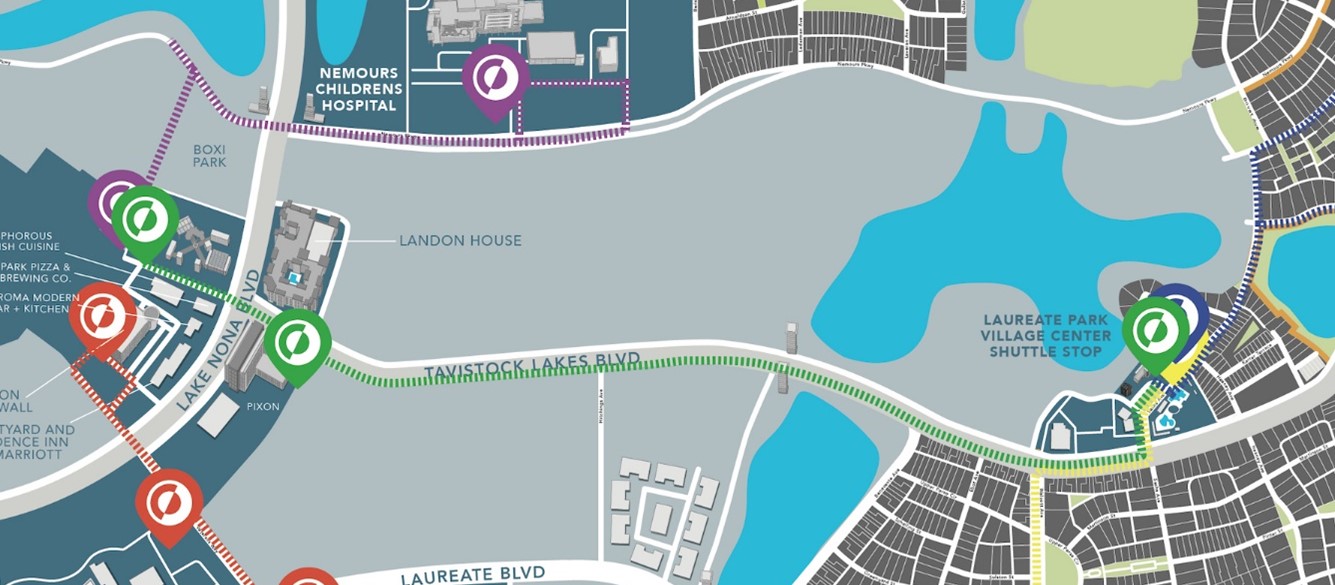}
  \caption{This figure shows in green the AS route in the Lake None city, from which it was collected the trajectories data. }\label{fig:route}
\end{figure}

The data was collected from autonomous shuttles (AS) manufactured by  NAVYA (Figure~\ref{fig:nona_av}). These vehicles have a passenger capacity of up to 15, are fully electric, and can operate for 9 hours per charge.  In this study, the vehicle operated in a residential area in Lake Nona, Florida. Data collection was part of an agreement between the company that coordinates AS operations in Lake Nona and the University of Florida.  The AS operates on five routes in Lake Nona. Route 1 was selected for this study because it operates on a public road with mixed traffic; it has a higher traffic volume than the other routes, and most of the route is on two-lane roads, which increases the likelihood of observing vehicles in car-following mode.  Route 1 (Figure~\ref{fig:route}) connects downtown Lake Nona, parks, and two neighborhoods. 

The total data collected for this study includes more than 20 AS' trips in both eastbound and westbound directions.  GPS collected data was used to extract the trajectory and a video was recorded to verify when the shuttle was running behind the conventional vehicle.

During the data collection, a conventional car traveled in front of the AS.  Raw trajectory data consists of GPS data location of one-second intervals from both vehicles. The data frequency was determined based on the previous study finding that the common sampling rates of 10 Hz are unnecessarily high and that 1 Hz suffices \cite{Treiber2013}. The longitude and latitude points were converted into distances in feet units, considering angular distances. To calculate the linear distances, the Geopy Python library was applied, and the distance between the points was calculated with the function \textit{geopy.distance}. The library applied is available on the website \textit{https://github.com/geopy/}. 

\subsection{Data cleaning}

The data collected resulted in a sample of 6,433 observations; for this study, each observation is a one-second time frame. After cleaning the data set and selecting only the time frame representing the car-following behavior, the data set included 4,427 observations.

The final data set includes the time frame when the AS is in car-following, and it does not include outliers. To clean the data set, we excluded data points in which the conventional vehicle was not in front of the AS or the AS was not moving (speed = 0 ft/s). Observations that resulted in a car or AV acceleration greater than 18 $ft/s^2$ or an AV speed greater than 15 $mi/h$ were considered outliers, as the vehicle is set to a maximum speed of 13 m/h. 

\subsection{Trajectory analyses}

The analyses of AS trajectories showed the histogram of the distribution of speed, acceleration, and jerk. The maximum acceleration and jerk observed on the trajectories were compared with the threshold proposed by other studies to assess the impact of the characteristics of the trajectories on user comfort\cite{deWinkel2023StandardsJerk}.

The Shapiro-Wilk test was applied to test the normality distribution for the variables, and Spearman’s correlation was calculated to investigate the correlation between speed, acceleration, jerk, spacing, and speed difference (delta speed).

The data set was divided into segments of trajectories when vehicles accelerate (acc+) and decelerate (acc-). The variance of speed, acceleration, and jerk was evaluated using the coefficient of variance (CV) and the mean percentage of outliers per trip \cite{Kummetha2023ExaminingTrajectories}. In this analysis, outliers were identified using the IQR measure of variability ($IQR = Q_3 - Q_1$).

To calculate the IQR, the data set was divided into quartiles. The time steps were sorted in ascending order and divided into 4 equal parts: $Q_1$ represents the $25^{th}$ percentile, $Q_2$ represents the $50^{th}$ percentile, and $Q_3$ represents the $75^{th}$ percentile. 
The IQR considers the range between the first and third quartiles, referred to as  $Q_1$ and $Q_3$. Data points below $Q_1 – 1.5 \times IQR$ or above $Q_3 + 1.5 \times IQR$ were considered outliers.

\subsection{Model calibration}

In this study, a genetic algorithm (GA) was applied to calibrate a car follower model for the following reasons:  (1)  GA has shown greater reliability for car-following model calibrations; (2) GA calibrations can capture correlations of speed and gaps \cite{Treiber2013MicroscopicApproach}, and (3) the random nature of GA influences the replicability of the best solution \cite{Punzo2012CanTrusted}. Several studies applied GA calibration and showed that GA is capable of identifying the appropriate values for calibrated model parameters \cite{Punzo2012CanTrusted, Punzo2021AboutCodes}. In addition to that, GA has been proven to be efficient in terms of convergence rate, resource usage, and quality of solutions to calibrate microscopic traffic models \cite{Punzo2012CanTrusted}.  

The calibration procedure is initialized by choosing a set of potential solutions, or individuals, which constitute a population. 
In this study, we used an initial population of 100 individuals, pseudorandomized within the constraints set for each parameter.  The evolutionary process is based on offspring selection, in which the preselected measure of performance (MOP)  scores individual members of the population. 

 The MOP provides a trade-off of performance, and it is used to evaluate the model prediction. Quantitative measures of the error between the observed and simulated trajectories were considered.

\begin{figure}[!ht]
  \centering
  \includegraphics[width=0.45\textwidth]{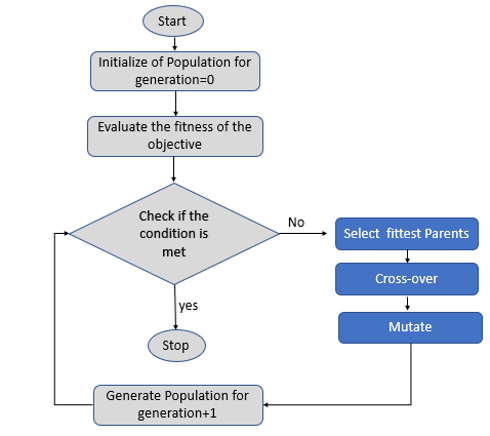}
  \caption{The process to calibrate the model with Genetic Algorithm (GA)}
  \label{fig:ga}
\end{figure}

We set the process to 1000 maximum interactions.  For each interaction, the individuals that resulted in the best calibration (which is the minimum error between the predicted and observed spacing) were selected for the next generation. The probability of individual mutation was established as 10\%, the crossover probability was 0.5, and the elitism ratio was 0.1 (that the best 10\% of individuals in a generation form the entire elite group and the best 50\% of individuals in the elite group directly propagate to the next generation). These parameters were set according to the Python library users' guide (https://github.com/PasaOpasen/geneticalgorithm2/blob/master) and were based on several tests and data sets to increase the chance of obtaining the best solution faster.  
 Thus, the individuals that result in the best MOP are selected as calibrated model parameters. This process is summarized in Figure \ref{fig:ga}. 

The open-source Distributed Evolutionary Algorithms in Python was used in this study. Thus, the GA can be easily replicated to develop car-following models for other vehicle types \cite{Hammit2018AData}.

The optimal solution was obtained considering the best fit of the parameters for the calibration data set. Each segment of car-following data was considered a trip.  Therefore, the following logic was applied: (i) obtain a set of calibration parameters; for GA it is known as the initial population; (ii) calculate the error considering all car-following segments in the calibration data; (iii)  identify the calibration parameter set that produces the minimum error.

The calibration used a Python library available on the website:
https://github.com/
PasaOpasen/geneticalgorithm2. It consists of a Python library to implement classic and elitist GA. This library has several advantages, such as allowing calibration with a very flexible structure, being faster than other libraries tested, and using pure Python.

Previous studies suggested that the use of spacing (following distance) for calibration results in an acceptable following speed; however, the inverse is not true \cite{Hammit2018AData, Kesting2008CalibratingData}. For this reason, after an extensive literature review, it was determined that the normalized root mean squared error (NRMSE) in the spacing is the MOP most suitable for this calibration \cite{Punzo2021AboutCodes}. The NRMSE is calculated according to Equation~\ref{eq:nerro} and Equation~\ref{eq:rmse}.

\begin{equation}
\operatorname{NRMSE}(x)=\frac{\operatorname{RMSE}(x)}{\sqrt{\frac{1}{T} \cdot \sum_{t^{-1}}^{t}\left[x^{\mathrm{obs}}(t)\right]^{2}}}
\label{eq:nerro}
\end{equation}

\begin{equation}
\operatorname{RMSE}(s)=\sqrt{\frac{1}{T} \cdot \sum_{t-1}^{T}\left[s^{s i n}(t)-s^{o b s}(t)\right]^{2}}
\label{eq:rmse}
\end{equation}

The car-following model was calibrated with 80\% of the trajectory data set and validated with 20\% of the remaining data. The validation data set was used to evaluate the accuracy of the models. The observed and simulated trajectories are evaluated using the God of Fitness: mean absolute error (MAE), root mean square error (RMSE), and normalized root mean square error (NRMSE). 

\subsection{Model comparison}

The IDM, ACC, and IIDM models were selected to calibrate the AS car-following model.

\subsubsection{IDM Car-following Model}

The IDM model is a deterministic model in which the acceleration of the vehicle depends on its own speed, the gap between vehicles, and the difference in speed.

\newenvironment{conditions*}
  {\par\vspace{\abovedisplayskip}\noindent
   \tabularx{\columnwidth}{>{$}l<{$} @{${}={}$} >{\raggedright\arraybackslash}X}}
  {\endtabularx\par\vspace{\belowdisplayskip}}
  
\begin{equation}
\begin{aligned}
& a_{I D M}(s, v, \Delta v)=a\left[1-\left(\frac{v}{v_0}\right)^\delta-\left(\frac{s^*(v, \Delta v)}{s}\right)^2\right] \\
& s^*(v, \Delta v)=s_0+\max \left(0, v T_0+\frac{v \Delta v}{2 \sqrt{a b}}\right)
\end{aligned}
\end{equation}

where:
\begin{conditions*}
 a    &  Maximum acceleration \\
\delta     &  Acceleration exponent \\
V_{0}     &  Desired speed \\
S_{0}     &  Desired standstill distance or Jam distance  \\
T_{0}     &  Desired time-gap \\
b     &  Desired deceleration  \\
\Delta v  & Speed difference between leading and following vehicle \\
s    & Space gap between leading and following vehicle.
\end{conditions*}

The model parameter $ \delta$ represents the acceleration reduction rate when the vehicle
nearly approaches the desired speed.
($ \delta$ = 1 corresponds
to a linear decrease while $ \delta$ → $\infty$ denotes a constant acceleration).
In the original study \cite{Treiber2000CongestedSimulations}, the $ \delta$ was 4, and many other studies have used 4 as a default value. When combined, the most simple functional form is given by:

Let \( a_I = a_{\mathrm{IDM}} \) and \( a_C = a_{\mathrm{CAH}} \). Then we have:
\begin{equation}
a_{\mathrm{ACC}}= 
\begin{cases}
a_I & \text{if } a_I \geq a_C, \\
(1-c) a_I+c\left[a_C+b \tanh \left(\frac{a_I-a_C}{b}\right)\right] & \text{otherwise.}
\end{cases}
\end{equation}
\\
The maximum acceleration $CAH$ that prevents crashes is given by:
\begin{equation}
\begin{aligned}
& a_{\mathrm{C}}\left(s, v, v_1, a_1\right) \\
& \quad= \begin{cases}\frac{v^2 \widetilde{a_l}}{v_1^2-2 s \tilde{a_l}} & \text { if } v_1\left(v-v_1\right) \leq-2 s \widetilde{a_l}, \\
\widetilde{a_l} \frac{\left(v-v_1\right)^2 \theta\left(v-v_1\right)}{2 s} & \text { otherwise, }\end{cases}
\end{aligned}
\end{equation}
\\
The improved IDM model (IIDM) \cite{Kesting2010EnhancedCapacity} contains one additional parameter compared to the IDM, the coolness factor, which reflects the model sensitivity with respect to changes in the gap. When c=1, the sensitivity of gap changes vanishes under small gaps, and no velocity difference exists. The original study assumed c =0.99.

The effective acceleration $\widetilde{a_l}$ has been used to avoid higher accelerations. The condition $v_1\left(v-v_1\right)=v_1 \Delta v \leq-2 s a_1$ is true if the vehicles have stopped at the time the minimum gap $s = 0$ is reached. Otherwise, negative approaching rates do not make sense and are eliminated by the Heaviside step function.

\subsubsection{ACC Car-following Model}

The model ACC model \cite{Milanes2014ModelingData} considers the acceleration of the vehicle based on distance and speed errors. 

\begin{equation}
a_{i, k}=k_1 \cdot e_{i, k}+k_2 \cdot\left(v_{i-1, k-1}-v_{i, k-1}\right)
\end{equation}
\begin{equation}
e_{i, k}=x_{i-1, k-1}-x_{i, k-1}-d_0-t_{\mathrm{des}} \cdot v_{i, k-1}
\end{equation}
where:
\begin{conditions*}
 a_k    &  Follower acceleration  \\
x_{k-1}     &  Leader position \\
x_{k}     &  Follower position \\
e_{i,k}     & Gap  error  of  vehicle  i  at  time  step  k  \\
v_{i,k-1}     &  Follower peed in the previous time step  \\
v_{i,k-1}     &  Leader peed in the previous time step  \\
d_0     &  Vehicle length in the gap error term  \\
t_{des}   & Desired time gap \\
s    & Space gap between leading and following vehicle.
\end{conditions*}

\section{RESULTS}

The descriptive statistics presented in Table~\ref{tab:descriptive} show that the AV travels at a maximum speed of 20 ft/s, the maximum speed set for AS operation, and a mean speed of 11.92 ft/s. The maximum acceleration was 5.29 $ft/s^2$, and the minimum was -17.69 $ft/s^2$. These values exceeded the acceleration thresholds of
 2.96 $ft/s^2$ for acceleration, and 1.97 $ft/s^3$ for a jerk, proposed to seat and standing AS passenger comfort \cite{Bae2019TowardBus}. 
 
 The data collected indicated that about 16\% of time travel resulted in jerks higher than the limit of excellent comfort ($0.92 ft/s^{3}$), 3.57\%  of travel time was higher than the excellent threshold ($4.03 ft/s^{3}$) and 2.24\% with jerk higher than expected ($4.82 ft/s^{3}$) \cite{deWinkel2023StandardsJerk}. These results suggest that adjusting the speed profile to reduce the peaks of jerk (usually caused by AS emergency stops), could improve the user's comfort when the AS is following another vehicle.

\begin{table}[htbp]
  \centering
  \caption{Descriptive statistics for AS trajectories}
    \begin{tabular}{| c | c | c | c | c |}
 \hline
    & \multirow{-1}{*}{Speed} & \multirow{-1}{*}{Acceleration} & \multirow{-1}{*}{Jerk} & \multirow{-1}{*}{Spacing} \\
    & ($feet/s$) & ($feet/s^2$) & ($feet/s^3$) & ($feet$) \\
    \hline
    mean  & 11.921 & -0.001 & 0.000 & 117.321 \\
    \hline
    std   & 6.109 & 1.515 & 2.106 & 126.835 \\
    \hline
    min   & 0.000 & -17.685 & -19.458 & 0.6900 \\
    \hline
    25\%  & 6.478 & -0.280 & -0.320 & 66.335 \\
    \hline
    50\%  & 14.390 & 0.010 & -0.010 & 97.050 \\
    \hline
    75\%  & 17.190 & 0.310 & 0.300 & 137.455 \\
    \hline
    max   & 20.470  & 5.290 & 19.476 & 1436.860 \\
    \hline
    \end{tabular}%
  \label{tab:descriptive}%
\end{table}%

\begin{figure*}[!t]
\centering
\subfloat[Speed histogram from AV trajectories]{\includegraphics[width=2.5in]{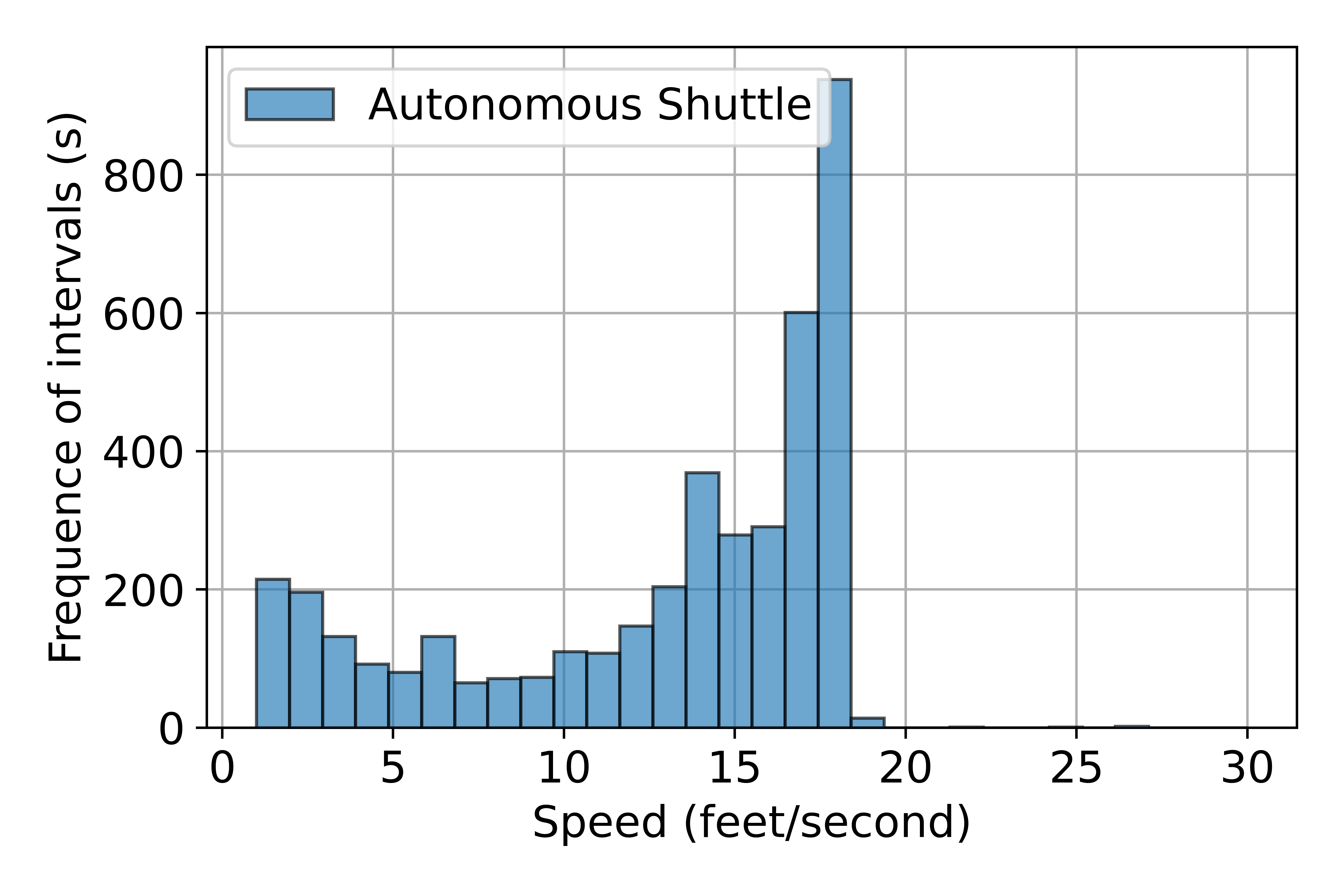}}
\hfil
\subfloat[Acceleration histogram from AV trajectories]{\includegraphics[width=2.5in]{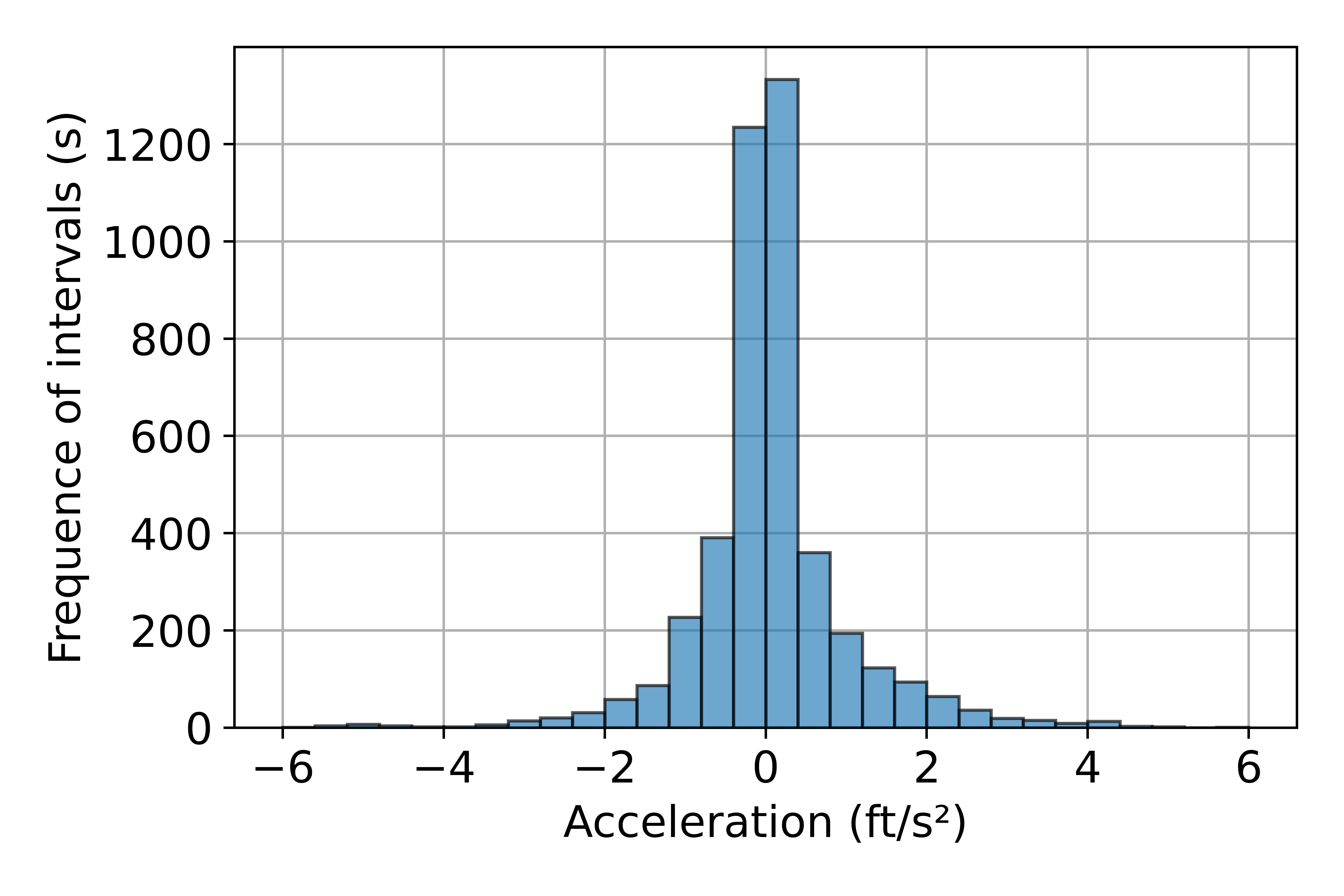}}
\\
\subfloat[Jerk histogram from AV trajectories]{\includegraphics[width=2.5in]{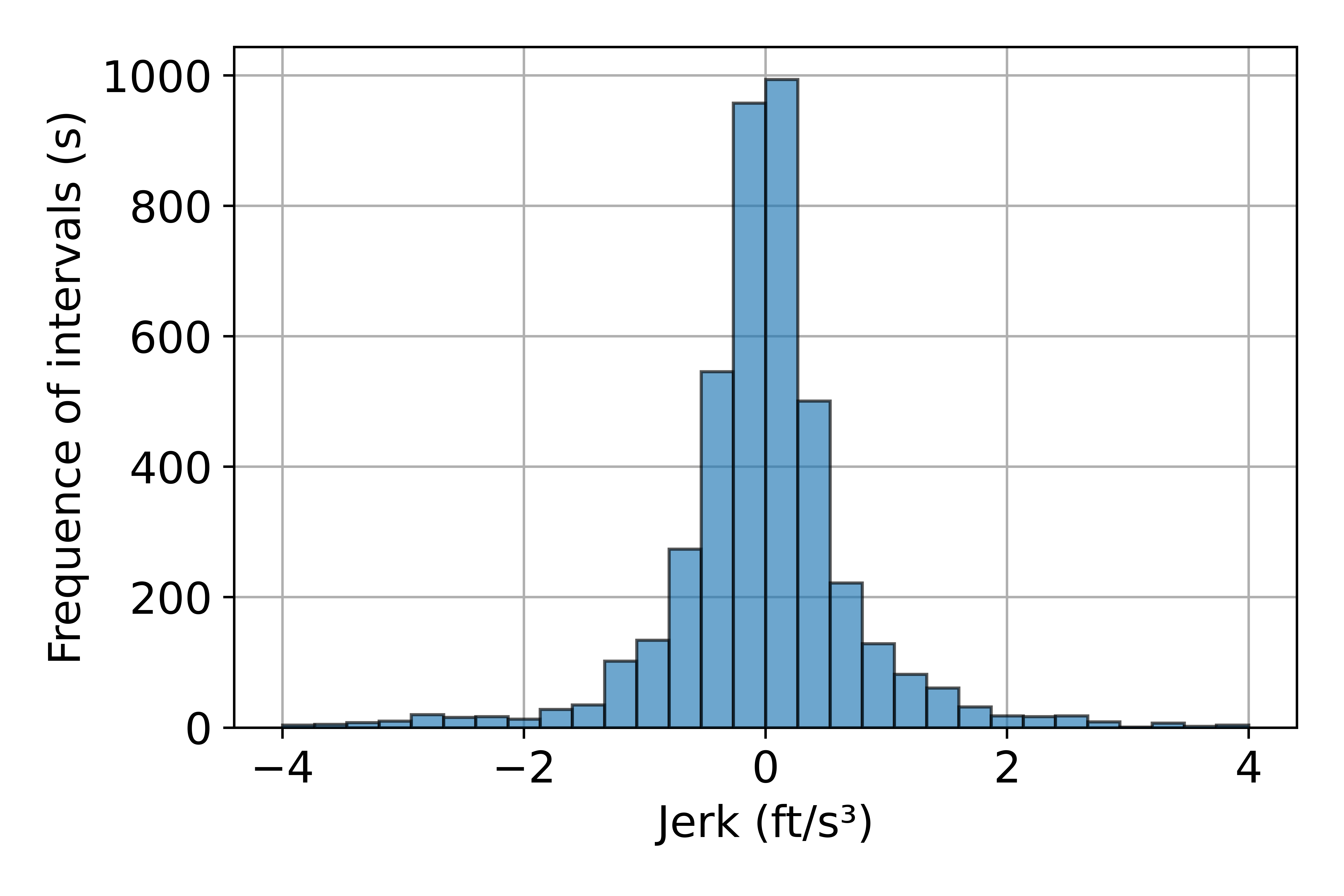}}
\hfil
\subfloat[Spacing between vehicles]{\includegraphics[width=2.5in]{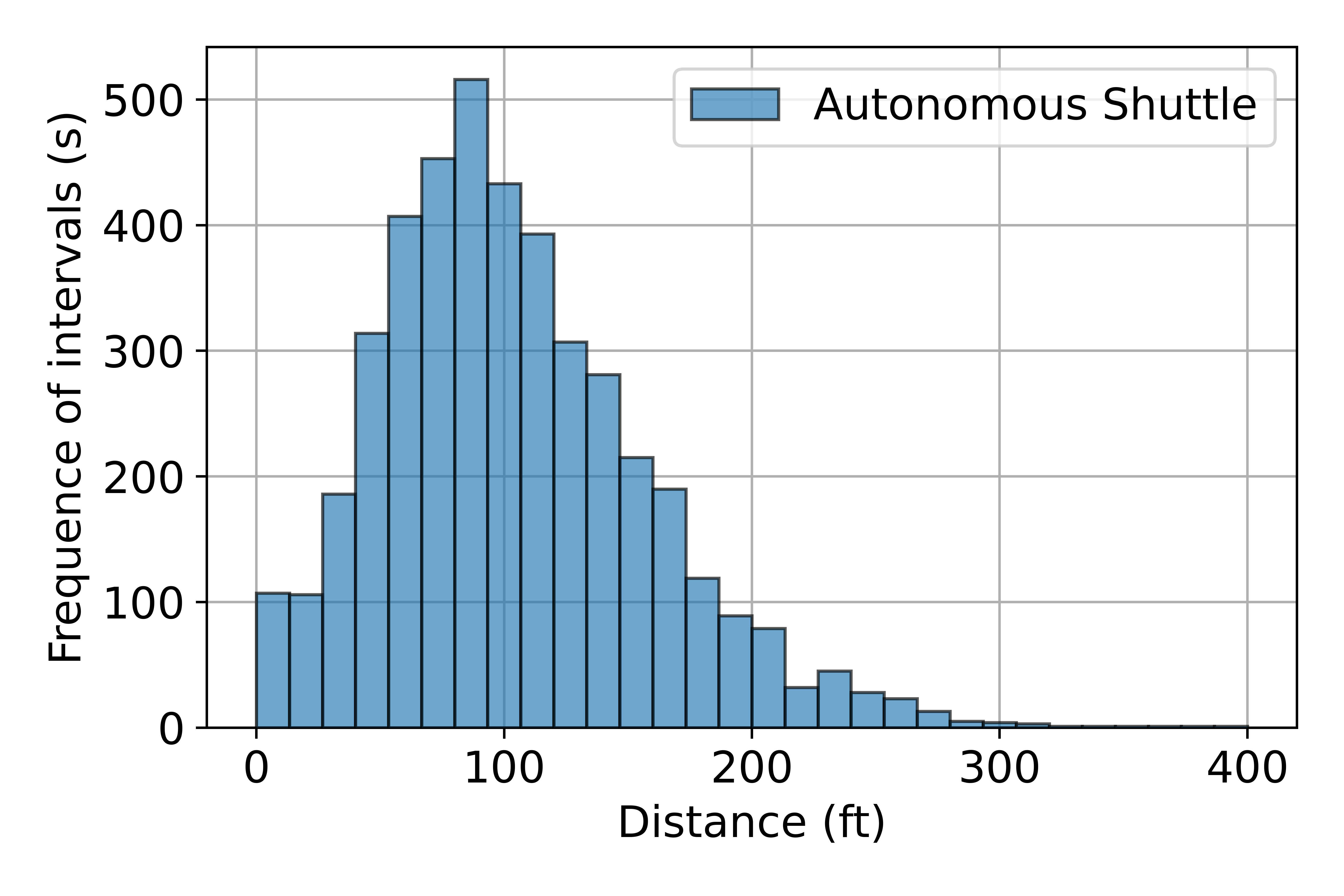}}
\hfil
\caption{Histogram of AV trajectory variables}
\label{fig:histograms}
\end{figure*}

The distributions for the variables speed, acceleration, jerk, and spacing between vehicles (Figure \ref{fig:histograms} a - c) as well as the result of the Shapiro-Wilk test suggested that the variables do not follow a normal distribution since the p-value is less than 0.05. The distribution of the variable "distance"(Figure \ref{fig:histograms}- d) indicated that the collected AS data represent the AS car-following mode since most of the time, the distance between vehicles was less than 300 ft, and less than the maximum range of AS sensors, which is 656 ft. \cite{Maehara2021CALIBRATIONSHUTTLE}.

\begin{figure*}[!t]
\centering
\subfloat[Speed variability]{\label{a}\includegraphics[width=2.7in]{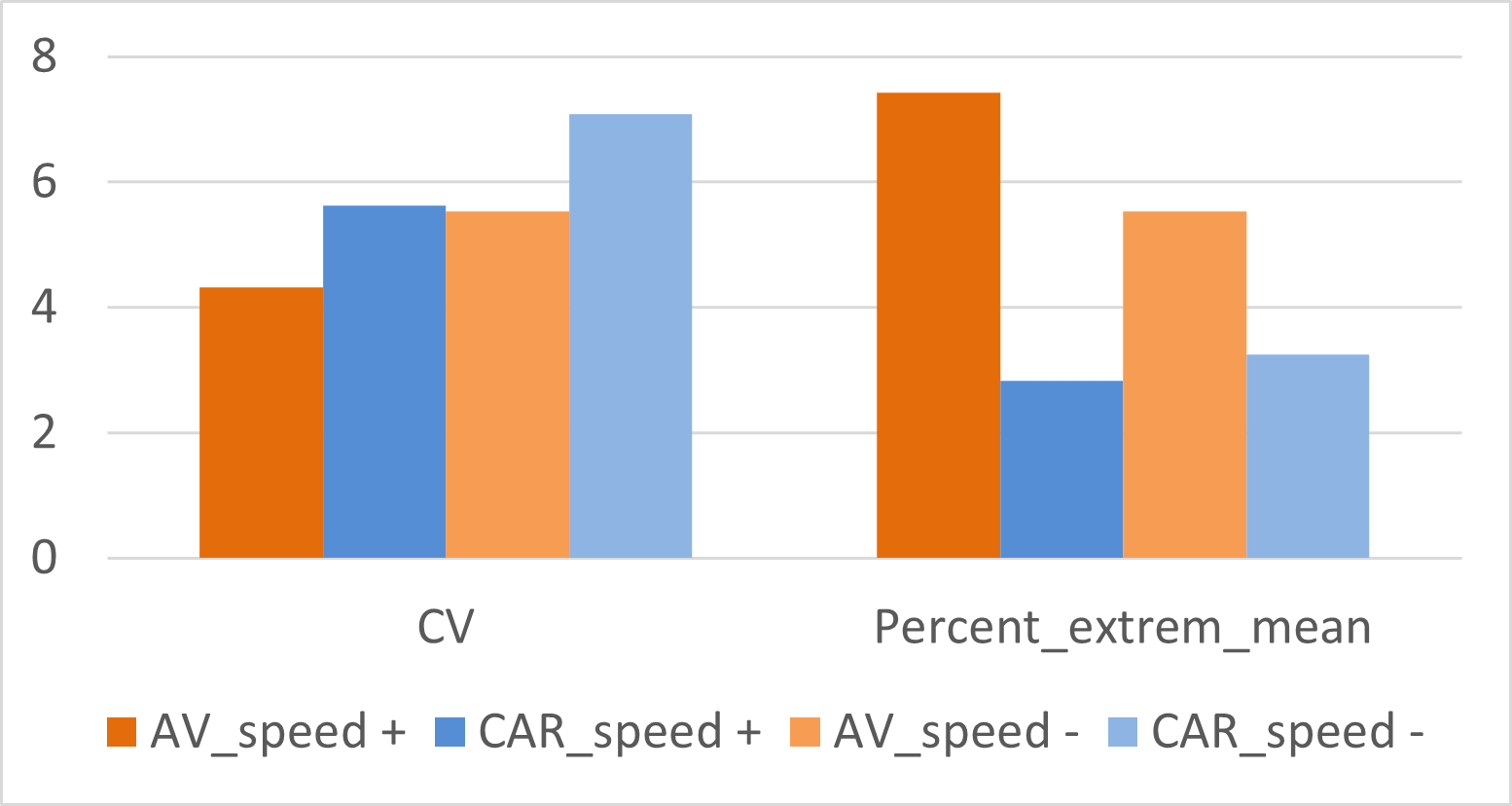}}
\hfill
\subfloat[Acceleration variability]{\label{b}\includegraphics[width=2.7in]{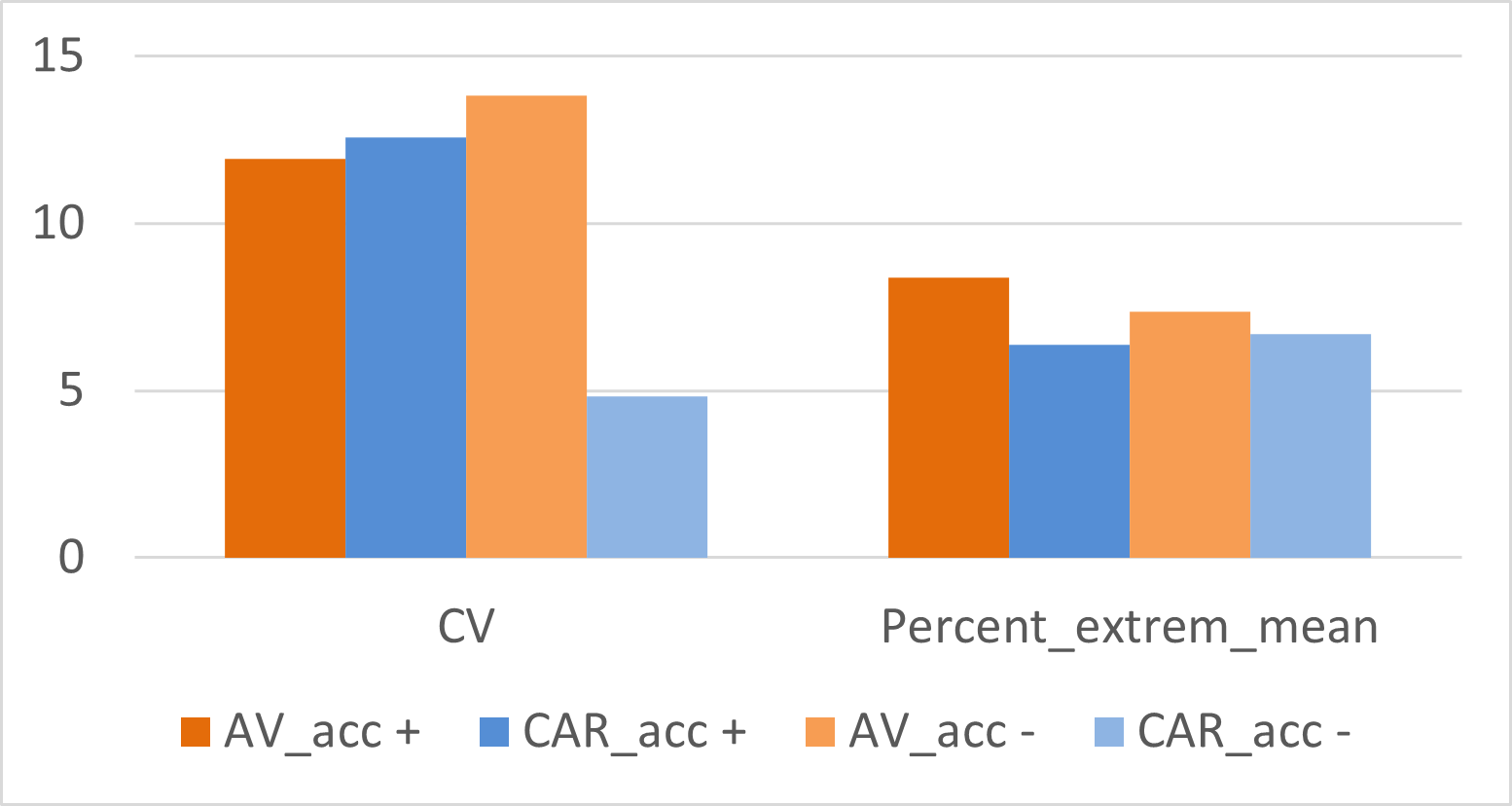}} 
\subfloat[Jerk variability]{\label{c}\includegraphics[width=2.7in]{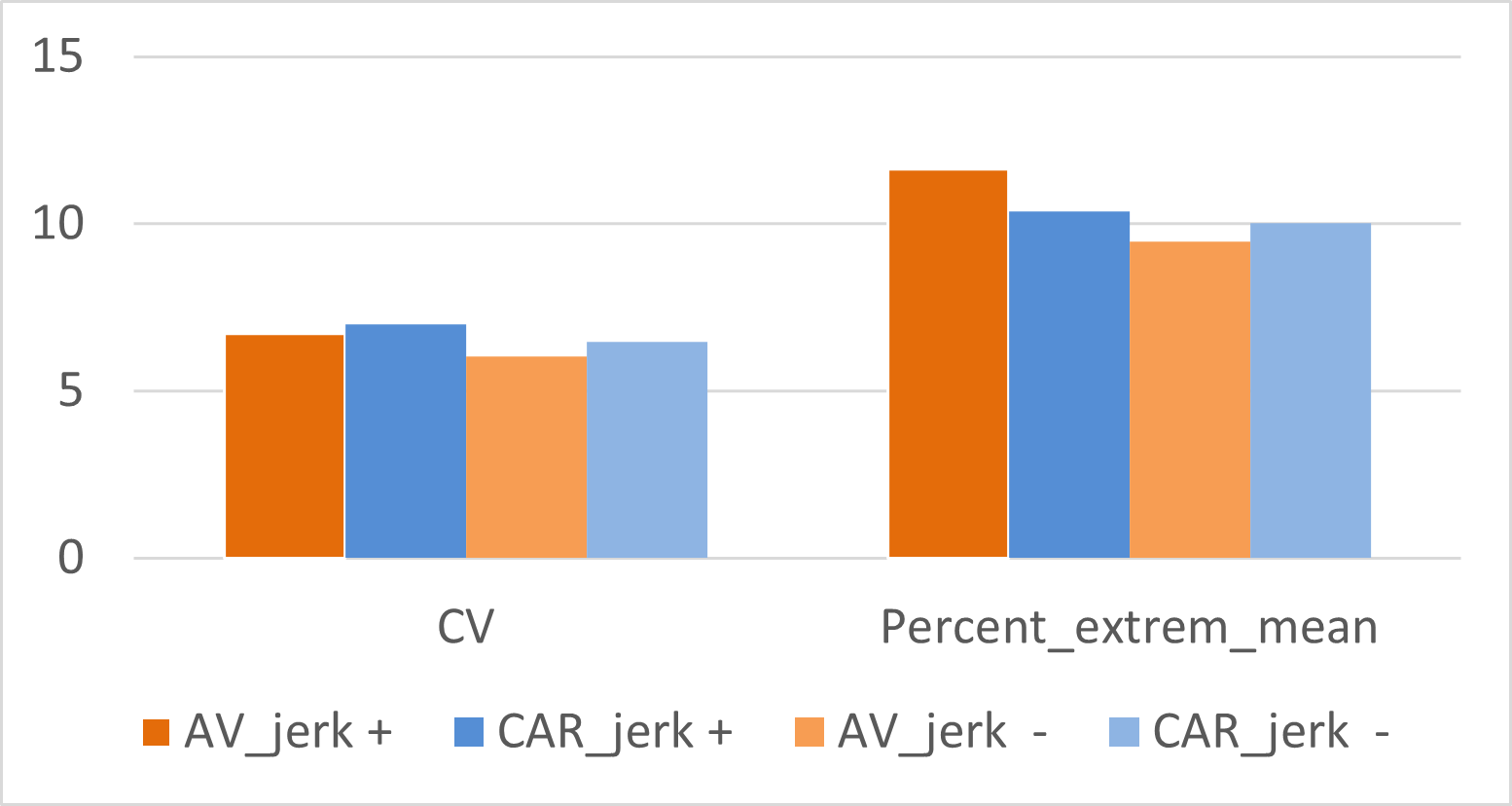}}
\caption{Speed, acceleration and jerk variability}
\label{volatil}
\end{figure*}

The variability measures (Figure(\ref{volatil}-(a)) calculated for the trajectories suggest that the follower vehicle (AV) exhibits less variance compared to the leader (car). This indicates that even if the leader has a higher level of speed dispersion, the AS could drive with a more uniform speed. However, AS trajectories had more outliers when considering the percentage of extreme values on the trips. This can be explained by the higher number of observations with an AS maximum speed (about 19 ft/s) and very low speed (less than 5 ft/s).

Figure \ref{volatil}-(b) presents a coefficient of variance (CV) for the variable $av_{acc+}$  lower than $av_{acc-}$, suggesting that the acceleration of the AV is more uniform than the deceleration of the AV. 

Figure \ref{volatil}-(c) shows a higher percentage of outliers for the AV jerk when the AV decelerates $av_{jerk-}$, which can be explained by the events of emergency stops of AS that occur when the AS applies a higher deceleration to stop immediately. These stops can be caused by the identification of obstacles on the mapped trajectory or by sensors' malfunction. In both cases, they directly affect passenger comfort.

\begin{figure}[!t]
  \centering
  \includegraphics[width=0.5\textwidth]{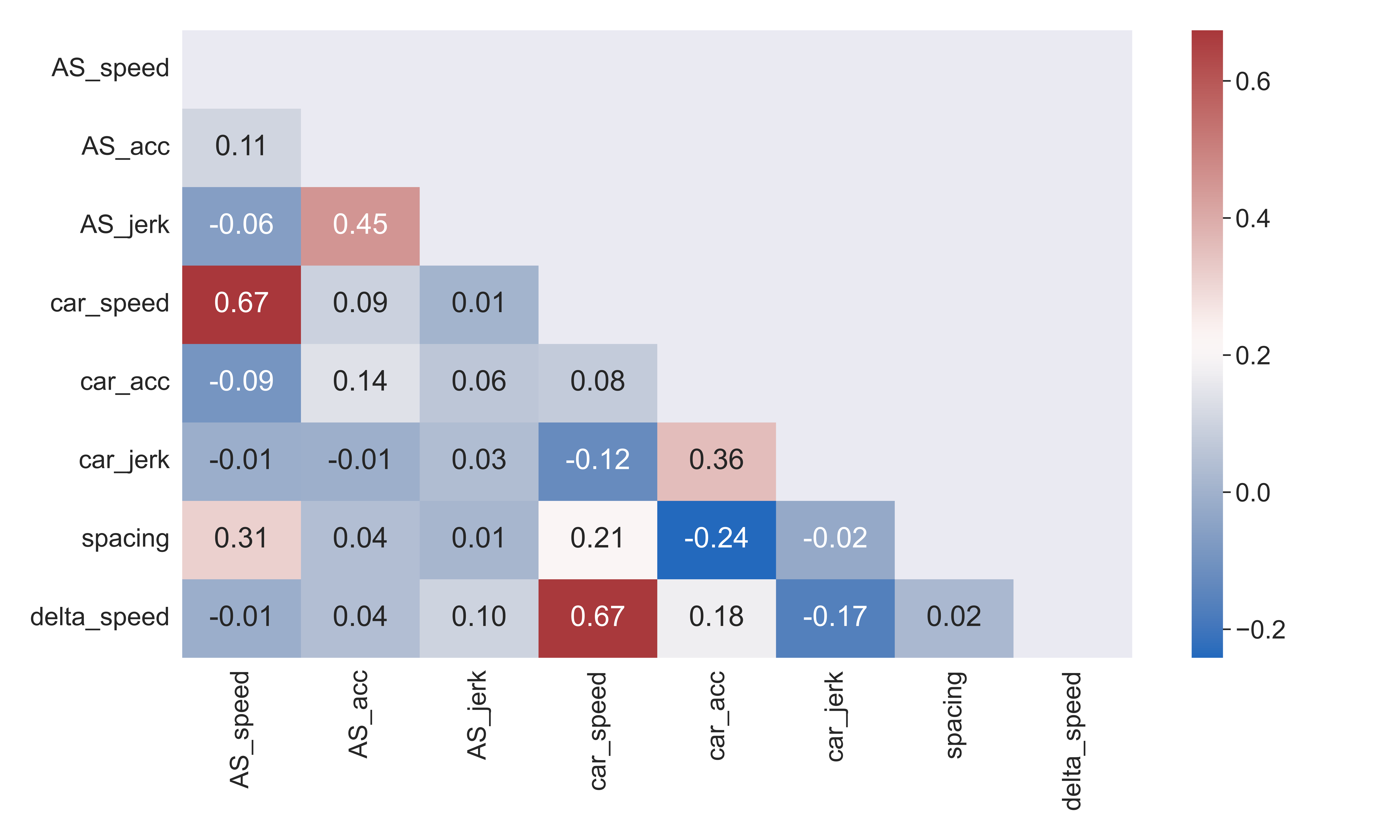}
  \caption{Spearman's correlation from the variables }
  \label{fig:correl}
\end{figure}

Since the variables are non-normal, the Spearman correlation was calculated to evaluate the relationship between the variables. Figure \ref{fig:correl} shows that the spacing between vehicles has a positive correlation with the speed of AS. The result suggests that vehicle spacing is an important variable to include in a car-following model to AS. 

\subsection{Models Calibrated}

A total of 10,000 calibrations were performed through 1,000 runs of GA with 10 seeds. The GA  was run on a Windows 10 computer with an Intel(R) Core(TM) i7-9700, CPU 3.00 GHz, and 32.0 GB RAM. The total computing time for the calibration was approximately 22 hours. The best results for each model were selected and validated with a separate data set. 
The acceleration and speed predicted by the models were limited by the maximum deceleration of -26 $ft/s^2$, the maximum speed at 19.5 $ft/s$, and the maximum acceleration of 10 $ft/s^2$ based on the values obtained in the field trajectories Table~\ref{tab:descriptive}.  In these calibrations, it was assumed that the AS had a reaction time of zero. 

The errors measured for the car-following calibration are shown in Table~\ref{tab:calibra_error}. Although the results suggest that, in general, the calibrated models simulate AS trajectories with a similar error measure, the ACC model performed slightly better. (Table~\ref{tab:valida_error}). This suggests that the ACC model can be more robust.  The ACC model also has the advantage of having fewer parameters to calibrate and, consequently, it can be calibrated and simulated faster than IDM and IIDM. The ACC model considers the difference in speed and spacing between vehicles and confirms that these variables are correlated with AS speed and acceleration (Table~\ref{fig:correl}).

\begin{table*}[!t]
  \centering
  \caption{Comparison of the errors estimated for the data set calibration}
    \begin{tabular}{| l | r | r | r | r | r | r |}
    \hline
          & \multicolumn{6}{c|}{Calibration } \\
          \hline
          & \multicolumn{3}{c|}{Spacing (feet)} & \multicolumn{3}{c|}{Speed (feet/second)} \\
          \hline
    Error & \multicolumn{1}{l}{IDM} & \multicolumn{1}{l}{ACC} & \multicolumn{1}{l|}{IIDM} & \multicolumn{1}{l}{IDM} & \multicolumn{1}{l}{ACC} & \multicolumn{1}{l|}{IIDM} \\
    \hline
    NRMSE & 0.01131187 & 0.01029776 & 0.01108331 & 0.17473816 & 0.18449651 & 0.17679173 \\
    \hline
    MAE   & 40.0936926 & 36.1561113 & 38.8674640 & 2.4326484 & 2.73698432 & 2.49166228 \\
    \hline
    RMSE  & 54.8555331 & 49.9566059 & 53.7674467 & 3.7603653 & 3.97036487 & 3.80455797 \\
    \hline
    \end{tabular}%
  \label{tab:calibra_error}%
\end{table*}%

\begin{table*}[!t]
  \centering
  \caption{Comparison of the errors estimated for the data set calibration}
    \begin{tabular}{| l | r | r | r | r | r | r |}
    \hline
          & \multicolumn{6}{c|}{Validation} \\
          \hline
          & \multicolumn{3}{c|}{Spacing (feet)} & \multicolumn{3}{c|}{Speed (feet/second)} \\
          \hline
    Error & \multicolumn{1}{l}{IDM} & \multicolumn{1}{l}{ACC} & \multicolumn{1}{l|}{IIDM} & \multicolumn{1}{l}{IDM} & \multicolumn{1}{l}{ACC} & \multicolumn{1}{l|}{IIDM} \\
    \hline
    NRMSE & 0.00976978 & 0.00883669 & 0.00945108 & 0.19100228 & 0.17900395 & 0.19403550 \\
    \hline
    MAE   & 37.2813140 & 29.8751013 & 35.3687157 & 2.32848233 & 2.26705052 & 2.34818250 \\
    \hline
    RMSE  & 43.9185694 & 39.7505274 & 42.5142538 & 3.45714136 & 3.23997141 & 3.51204262 \\
    \hline
    \end{tabular}%
  \label{tab:valida_error}%
\end{table*}%

The parameters calibrated for the model IDM, ACC, and IIDM are shown in Table~\ref{tab:idm}, Table~\ref{tab:acc}, and Table~\ref{tab:iidm}, respectively. Models calibrated for AS resulted in lower maximum acceleration and higher deceleration compared to the parameters suggested for other vehicles in the literature.
In the IDM model, the acceleration exponent is equal to 1, indicating that the acceleration is nearly linear, which is expected for electric vehicles. The distance between the jam and the desired time gap indicates that the AS is more cautious and maintains a longer distance between the AS and the vehicle leader. The desired deceleration is four times higher than that estimated by other authors.
In the ACC mode, the calibrated time gap is also higher than the values estimated in the study  \cite{Milanes2014ModelingData}.
The parameters calibrated for the IIDM models indicated results similar to those of the IDM model. The value of $c$ factor less than 0.99 suggests that the AS trajectory is closer to the ACC model than the one estimated to AV \cite{Kesting2010EnhancedCapacity} .

\begin{table*}
\footnotesize
\centering
  \caption{Parameters calibrated for IDM}
    \begin{tabular}{|c|l|r|r|r|r|r|r|}
    \hline
    \multicolumn{1}{|r}{} & \multicolumn{1}{|c|}{Description} & AS &  Kesting \textit{et. al} \cite{Kesting2008AdaptiveAvoidance} &  Zhu \textit{et. al} \cite{Zhu2018ModelingStudy} &  Treiber \textit{et. al} \cite{Treiber2000CongestedSimulations} & Kovacs \textit{et. al}\cite{Kovacs2016ParametriKrizanjima} & Salles \textit{et. al}\cite{Salles2022ExtendingMeasurements} \\
    \hline
    a     & Max acceleration $(ft/s^{2})$ & 2.76  & 4.59  & [0.33-16.40] & 2.40  & 5.25  & 8.20 - 17.39 \\
    \hline
    exp   & Acceleration exponent & 1     & 4     & [1-40] & 4     & 4     & 2 \\
    \hline
    $v_0$    & Desired speed $(ft/^s)$ & 20.00 & 100.25 & [0.09 - 136.70] & 109.36 & 50.12 & 45.57 \\
    \hline
    S0    & Jam distance $(ft)$ & 9.89  & 6.56  & [1.64 - 32.81] & 6.56  & 6.56  & [4.92 - 8.20]  \\
    \hline
    T     & Desired time gap (s) & 2.79  & 0.60  & [0.1 - 0.5] & 1.60  & 0.86  & [1.1-1.3] \\
    \hline
    b     & Desired deceleration $(ft/s^2)$ & 24.58 & 6.56  & [0.33 - 16.40] & 5.48  & 6.56  & [8.20 - 17.39]  \\
    \hline
\end{tabular}
\label{tab:idm}%
\end{table*}

\begin{table}[!t]
\small
  \centering
  \caption{Parameters calibrated for the ACC model}
    \begin{tabular}{|l|r|r|r|l|}
    \hline
          \multicolumn{1}{|r|}{Parameter}& \multicolumn{1}{l|}{AS} & \multicolumn{1}{l|}{Milanes \textit{et. al} \cite{Milanes2014ModelingData}} &  \multicolumn{1}{l|}{Xiao \textit{et. al}\cite{Xiao2017RealisticVehicles}  }\\
    \hline
    $t_{hw} (s)$ & 4.96  & [0.4-0.7] & [1.0 - 9.0] \\
    \hline
    $K_1 s^{-2}$ & 0.01  & 0.23  & 0.23  \\
    \hline
    $K_2 s^{-1}$ & 0.43  & 0.04  & 0.07   \\
    \hline
    \end{tabular}%
\label{tab:acc}%
\end{table}%

\begin{table}[!t]
\footnotesize
  \centering
  \caption{IIDM Parameters}
    \begin{tabular}{|c|l|r|r|}
     \hline
    \multicolumn{1}{|l|}{Parameter} & Description & \multicolumn{1}{l|}{AS} & \multicolumn{1}{l|}{ Kesting \textit{et. al} \cite{Kesting2010EnhancedCapacity}}   \\
     \hline
    a     & Max acceleration ($ft/s^2$) & 1.214 & 4.59   \\
     \hline
    exp   & Acceleration exponent & 3     & 4      \\
     \hline
    v0    & Desired speed $(ft/s)$ & 18.742 & 109.36  \\
     \hline
    S0    & Jam distance {($feet$)} & 9.892 & 6.562 \\
     \hline
    T     & Desired time gap ({$s$}) & 2.980 & 1.5   \\
     \hline
    b     & Desired deceleration ({$ft/s^2$})  & 24.846 & 6.56  \\
     \hline
    c     & Coolness factor  & 0.959 & 0.99   \\
     \hline
    \end{tabular}%
  \label{tab:iidm}%
\end{table}%

\section{Conclusions}

AS are being tested in several cities, and previous studies suggest that AS can impact the public transport network. However, there is a lack of simulation tools and car-following models that allow us to assess the impacts of AS before their implementation in the field. Furthermore, there are no studies that investigate AS trajectories with a significant sample size to calibrate car-following behavior. 
To fill these gaps, this project collected AS field data trajectories, assessed  AS trajectories' characteristics, calibrated AS car-following models using GA, and compared the calibrated models with parameters obtained for passenger AV in the literature.

 AS trajectories were collected with GPS equipment in the leader vehicle (conventional vehicle with a human driver) and in the follower (AS).  GPS positions were used to calculate the speed, acceleration, and jerk of vehicles. The data collected indicated that about 16\% of the time, travel presented jerks higher than the limit for excellent comfort. 
The acceleration of the AS is more constant than the deceleration of the AS, but shows a higher percentage of outliers for the AS jerk when the AS decelerates, very likely caused by AS emergency breaks. 

We selected the IDM, ACC, and IIDM to simulate AS trajectories because they are flexible, they were implemented to simulate AV and connected vehicles in other studies, and they could be calibrated for AS.  GA was selected to perform the calibration according to the results obtained in previous studies \cite{Punzo2021AboutCodes}. 

The ACC-calibrated model resulted in the best performance in simulating AS speed and spacing compared to IDM and IIDM.
The calibrated parameters suggest that, in general, the AS has lower maximum acceleration and higher deceleration compared to the parameters suggested in the literature. In the IDM model, the acceleration exponent is equal to 1 which indicates that the acceleration is close to linear. The jam distance and the desired time gap
indicate that the AS exhibits a more cautious behavior compared to the parameters estimated for AV in other studies. The calibrated parameters for the IIDM models indicated
results similar to those of the IDM model, and the c factor less than 0.99 suggests that the AV trajectory may be adequately simulated with an ACC model.

 This study represents an advancement  in the field of simulation of new transport technologies and will allow the evaluation of the implementation of AS under different conditions before AS field operation. It can improve AS-related impact studies, reducing the time and cost of pilot studies, and it can contribute to improving the public transport system. 

This study was limited to evaluating existing AS car-following models. For future studies, we suggest evaluating different models of AS,  AS's gap acceptance, autonomous shuttle V2I communication in intersections, and the lane-changing behavior of vehicles around the AS.



\section*{Acknowledgements}    
The research was funded by the US DOT, under Regional UTC STRIDE Project "Utilization of Connectivity and Automation in Support of Transportation Agencies’ Decision Making – Phase 2". The data collection was supported by Beep company.

\bibliography{IEEEabrv,references.bib}{}

\begin{thebibliography}{10}
\providecommand{\url}[1]{#1}
\csname url@samestyle\endcsname
\providecommand{\newblock}{\relax}
\providecommand{\bibinfo}[2]{#2}
\providecommand{\BIBentrySTDinterwordspacing}{\spaceskip=0pt\relax}
\providecommand{\BIBentryALTinterwordstretchfactor}{4}
\providecommand{\BIBentryALTinterwordspacing}{\spaceskip=\fontdimen2\font plus
\BIBentryALTinterwordstretchfactor\fontdimen3\font minus \fontdimen4\font\relax}
\providecommand{\BIBforeignlanguage}[2]{{%
\expandafter\ifx\csname l@#1\endcsname\relax
\typeout{** WARNING: IEEEtran.bst: No hyphenation pattern has been}%
\typeout{** loaded for the language `#1'. Using the pattern for}%
\typeout{** the default language instead.}%
\else
\language=\csname l@#1\endcsname
\fi
#2}}
\providecommand{\BIBdecl}{\relax}
\BIBdecl

\bibitem{sae2018}
{SAE}, ``{Taxonomy and definitions for terms related to driving automation systems for on-road motor vehicles},'' \emph{SAE Int.}, vol. 4970, pp. 1--5, 2018.

\bibitem{Iclodean2020AutonomousReview}
\BIBentryALTinterwordspacing
C.~Iclodean, N.~Cordos, and B.~O. Varga, ``{Autonomous Shuttle Bus for Public Transportation: A Review},'' \emph{Energies 2020, Vol. 13, Page 2917}, vol.~13, no.~11, p. 2917, 6 2020. [Online]. Available: \url{https://www.mdpi.com/1996-1073/13/11/2917/htm https://www.mdpi.com/1996-1073/13/11/2917}
\BIBentrySTDinterwordspacing

\bibitem{Feys2020ExperienceRegion}
M.~Feys, E.~Rombaut, and L.~Vanhaverbeke, ``{Experience and acceptance of autonomous shuttles in the brussels capital region},'' \emph{Sustainability (Switzerland)}, vol.~12, no.~20, pp. 1--23, 10 2020.

\bibitem{Piatkowski2021AutonomousThem}
\BIBentryALTinterwordspacing
D.~P. Piatkowski, ``{Autonomous Shuttles: What Do Users Expect and How Will They Use Them?}'' \emph{Journal of Urban Technology}, vol.~28, no. 3-4, pp. 97--115, 10 2021. [Online]. Available: \url{https://www.tandfonline.com/doi/full/10.1080/10630732.2021.1896345}
\BIBentrySTDinterwordspacing

\bibitem{Haque2020}
A.~M. Haque and C.~Brakewood, ``{A synthesis and comparison of American automated shuttle pilot projects},'' \emph{Case Studies on Transport Policy}, vol.~8, no.~3, pp. 928--937, 9 2020.

\bibitem{Calvert2017WillFlow}
\BIBentryALTinterwordspacing
S.~C. Calvert, W.~J. Schakel, and J.~W.~C. van Lint, ``{Will Automated Vehicles Negatively Impact Traffic Flow?}'' \emph{Journal of Advanced Transportation}, vol. 2017, pp. 1--17, 2017. [Online]. Available: \url{https://www.hindawi.com/journals/jat/2017/3082781/}
\BIBentrySTDinterwordspacing

\bibitem{Papadoulis2019EvaluatingMotorways}
A.~Papadoulis, M.~Quddus, and M.~Imprialou, ``{Evaluating the safety impact of connected and autonomous vehicles on motorways},'' \emph{Accident Analysis and Prevention}, vol. 124, pp. 12--22, 3 2019.

\bibitem{Oikonomou2020}
\BIBentryALTinterwordspacing
M.~G. Oikonomou, F.~P. Orfanou, E.~I. Vlahogianni, and G.~Yannis, ``{Impacts of Autonomous Shuttle Services on Traffic, Safety and Environment for Future Mobility Scenarios},'' in \emph{2020 IEEE 23rd International Conference on Intelligent Transportation Systems (ITSC)}.\hskip 1em plus 0.5em minus 0.4em\relax IEEE, 9 2020, pp. 1--6. [Online]. Available: \url{https://ieeexplore.ieee.org/document/9294576/}
\BIBentrySTDinterwordspacing

\bibitem{Lu2019Energy-EfficientVehicles}
C.~Lu, J.~Dong, and L.~Hu, ``{Energy-Efficient Adaptive Cruise Control for Electric Connected and Autonomous Vehicles},'' \emph{IEEE Intelligent Transportation Systems Magazine}, vol.~11, no.~3, pp. 42--55, 9 2019.

\bibitem{Liu2019CanEvaluation}
F.~Liu, F.~Zhao, Z.~Liu, and H.~Hao, ``{Can autonomous vehicle reduce greenhouse gas emissions? A country-level evaluation},'' \emph{Energy Policy}, vol. 132, pp. 462--473, 9 2019.

\bibitem{Alkim2007FieldDriver}
T.~P. Alkim, G.~Bootsma, and S.~Hoogendoorn, ``{Field operational test - The assisted driver},'' \emph{IEEE Intelligent Vehicles Symposium}, pp. 1198--1203, 2007.

\bibitem{Ainsalu2018StateBuses}
\BIBentryALTinterwordspacing
J.~Ainsalu, V.~Arffman, M.~Bellone, M.~Ellner, T.~Haapam{\"{a}}ki, N.~Haavisto, E.~Josefson, A.~Ismailogullari, B.~Lee, O.~Madland, R.~Mad{\v{z}}ulis, J.~M{\"{u}}{\"{u}}r, S.~M{\"{a}}kinen, V.~Nousiainen, E.~Pilli-Sihvola, E.~Rutanen, S.~Sahala, B.~Sch{\o}nfeldt, P.~M. Smolnicki, R.~M. Soe, J.~S{\"{a}}{\"{a}}ski, M.~Szyma{\'{n}}ska, I.~Vaskinn, and M.~{\AA}man, ``{State of the Art of Automated Buses},'' \emph{Sustainability 2018, Vol. 10, Page 3118}, vol.~10, no.~9, p. 3118, 8 2018. [Online]. Available: \url{https://www.mdpi.com/2071-1050/10/9/3118/htm https://www.mdpi.com/2071-1050/10/9/3118}
\BIBentrySTDinterwordspacing

\bibitem{Azad2019FullyDirections}
M.~Azad, N.~Hoseinzadeh, C.~Brakewood, C.~R. Cherry, and L.~D. Han, ``{Fully Autonomous Buses: A Literature Review and Future Research Directions},'' \emph{Journal of Advanced Transportation}, vol. 2019, 2019.

\bibitem{Li2020DataCities}
\BIBentryALTinterwordspacing
X.~Li, S.~Cheng, Z.~Lv, H.~Song, T.~Jia, and N.~Lu, ``{Data analytics of urban fabric metrics for smart cities},'' \emph{Future Generation Computer Systems}, vol. 107, pp. 871--882, 6 2020. [Online]. Available: \url{https://dl.acm.org/doi/abs/10.1016/j.future.2018.02.017}
\BIBentrySTDinterwordspacing

\bibitem{Elefteriadou2014ModelingVehicles}
\BIBentryALTinterwordspacing
L.~Elefteriadou, ``{Modeling Vehicle Interactions and the Movement of Groups of Vehicles},'' 2014, pp. 31--58. [Online]. Available: \url{http://link.springer.com/10.1007/978-1-4614-8435-6_2}
\BIBentrySTDinterwordspacing

\bibitem{Shi2021PreliminaryShuttle}
\BIBentryALTinterwordspacing
Y.~Shi, A.~P. Bartlett, R.~Dmowski, D.~Duchscherer, Q.~He, C.~Qiao, and A.~W. Sadek, ``{Preliminary Safety Evaluation of Self-Driving, Low-Speed Shuttle},'' \emph{Journal of Transportation Engineering, Part A: Systems}, vol. 147, no.~8, p. JTEPBS.0000535, 8 2021. [Online]. Available: \url{http://ascelibrary.org/doi/10.1061/JTEPBS.0000535}
\BIBentrySTDinterwordspacing

\bibitem{MahmoodiNesheli2021DriverlessDeployment}
M.~Mahmoodi~Nesheli, L.~Li, M.~Palm, and A.~Shalaby, ``{Driverless shuttle pilots: Lessons for automated transit technology deployment},'' \emph{Case Studies on Transport Policy}, vol.~9, no.~2, pp. 723--742, 6 2021.

\bibitem{Zhao2020FieldVehicle}
X.~Zhao, Z.~Wang, Z.~Xu, Y.~Wang, X.~Li, and X.~Qu, ``{Field experiments on longitudinal characteristics of human driver behavior following an autonomous vehicle},'' \emph{Transportation Research Part C: Emerging Technologies}, vol. 114, pp. 205--224, 5 2020.

\bibitem{Ahmed2019EvaluationVehicles}
S.~Ahmed, K.~Dey, and R.~Fries, ``{Evaluation of transportation system resilience in the presence of connected and automated vehicles},'' \emph{TRR: Journal of the TRB}, 2019.

\bibitem{DfT2016ResearchFlow}
{DfT}, \emph{{Research on the impacts of connected and autonomous vehicles (CAVs) on traffic flow}}.\hskip 1em plus 0.5em minus 0.4em\relax Reino Unido: ATKINS, 2016.

\bibitem{Morando2018StudyingMeasures}
M.~M. Morando, Q.~Tian, L.~T. Truong, and H.~L. Vu, ``{Studying the safety impact of autonomous vehicles using simulation-based surrogate safety measures},'' \emph{Journal of Advanced Transportation}, p.~11, 2018.

\bibitem{Stanek2018MeasuringSimulation}
D.~Stanek, E.~Huang, R.~T. Milam, and Y.~A. Wang, ``{Measuring Autonomous Vehicle Impacts on Congested Networks Using Simulation},'' in \emph{97th Annual Meeting of Transportation Research Board}.\hskip 1em plus 0.5em minus 0.4em\relax Washington: TRB, 2018, p.~16.

\bibitem{Sanusi2019MeasuringZones}
F.~B. Sanusi, J.~O. Sobanjo, E.~E. Ozguven, T.~Sando, and Y.~AbdelRazig, ``{Measuring impacts of connected and autonomous vehicles on freeway work zones},'' in \emph{98th Annual Meeting of Transportation Research Board}.\hskip 1em plus 0.5em minus 0.4em\relax Washington: TRB, 2019.

\bibitem{SenturkBerktas2020EffectIntersections}
E.~{\c{S}}ent{\"{u}}rk~Berkta{\c{s}} and S.~Tanyel, ``{Effect of Autonomous Vehicles on Performance of Signalized Intersections},'' \emph{Journal of Transportation Engineering, Part A: Systems}, vol. 146, no.~2, p. 04019061, 2 2020.

\bibitem{Shladover2012ImpactsFlow}
S.~E. Shladover, D.~Su, and X.~Lu, ``{Impacts of cooperative adaptive cruise control on freeway traffic flow},'' \emph{TRR: Journal of the TRB}, vol. 2324, pp. 63--70, 2012.

\bibitem{Hamilton2018DedicatingVehicles}
\BIBentryALTinterwordspacing
B.~A. Hamilton, \emph{{Dedicating Lanes for Priority or Exclusive Use by Connected and Automated Vehicles}}.\hskip 1em plus 0.5em minus 0.4em\relax Washington, D.C.: Transportation Research Board, 1 2018. [Online]. Available: \url{https://www.nap.edu/catalog/25366}
\BIBentrySTDinterwordspacing

\bibitem{Shi2019CapacityMethod}
Y.~Shi, Q.~He, and Z.~Huang, ``{Capacity Analysis and Cooperative Lane Changing for Connected and Automated Vehicles: Entropy-Based Assessment Method},'' \emph{TRR: Journal of the TRB}, vol. 2673, no.~8, pp. 1--14, 2019.

\bibitem{Sukennik2018Micro-simulationD2.5}
P.~Sukennik, \emph{{Micro-simulation guide for automated vehicles - D2.5}}.\hskip 1em plus 0.5em minus 0.4em\relax European Union: CoExist, 2018.

\bibitem{Hoogendoorn2014AutomatedReview}
R.~Hoogendoorn, B.~Van~Arem, and S.~Hoogendoorn, ``{Automated driving, traffic flow efficiency and human factors: a literature review},'' \emph{TRR: Journal of the TRB}, vol. 2422, pp. 113--120, 2014.

\bibitem{Treiber2013MicroscopicApproach}
M.~Treiber and A.~Kesting, ``{Microscopic Calibration and Validation of Car-Following Models: A Systematic Approach},'' \emph{Procedia - Social and Behavioral Sciences}, vol.~80, pp. 922--939, 6 2013.

\bibitem{Li2019ParsimoniousTraffic}
L.~Li and X.~Li, ``{Parsimonious trajectory design of connected automated traffic},'' \emph{Transportation Research Part B: Methodological}, vol. 119, pp. 1--21, 1 2019.

\bibitem{Roncoli2015ModelVehicles}
C.~Roncoli, I.~Papamichail, and M.~Papageorgiou, ``{Model Predictive Control for Motorway Traffic with Mixed Manual and VACS-equipped Vehicles},'' \emph{Transportation Research Procedia}, vol.~10, pp. 452--461, 2015.

\bibitem{Bao2020}
\BIBentryALTinterwordspacing
L.~Bao, Q.~Wang, G.~Xiao, and A.~Ni, ``{Determining the Influence of Auto-Driving on Urban Road Traffic Conditions},'' \emph{Resilience and Sustainable Transportation Systems - Selected Papers from the 13th Asia Pacific Transportation Development Conference}, pp. 734--745, 2020. [Online]. Available: \url{https://ascelibrary.org/doi/10.1061/9780784482902.086}
\BIBentrySTDinterwordspacing

\bibitem{Gasper2018SimulationSpace}
R.~Gasper, S.~Beutelschie{\ss}, M.~Krumnow, L.~Simon, Z.~Baksa, and J.~Schwarzer, ``{Simulation of Autonomous RoboShuttles in Shared Space},'' vol.~2, pp. 183--171, 6 2018.

\bibitem{He2021DesignBuses}
H.~He, M.~Shi, J.~Li, J.~Cao, and M.~Han, ``{Design and experiential test of a model predictive path following control with adaptive preview for autonomous buses},'' \emph{Mechanical Systems and Signal Processing}, vol. 157, 8 2021.

\bibitem{Li2021DevelopmentEnvironment}
\BIBentryALTinterwordspacing
X.~Li, S.~Zhu, B.~Aksun-Guvenc, and L.~Guvenc, ``{Development and Evaluation of Path and Speed Profile Planning and Tracking Control for an Autonomous Shuttle Using a Realistic, Virtual Simulation Environment},'' \emph{Journal of Intelligent and Robotic Systems}, vol. 101:42, 2021. [Online]. Available: \url{https://doi.org/10.1007/s10846-021-01316-2}
\BIBentrySTDinterwordspacing

\bibitem{Weissensteiner2021VirtualShuttles}
\BIBentryALTinterwordspacing
P.~Weissensteiner, G.~Stettinger, K.~Tieber, and K.~Rehrl, ``{Virtual Risk Assessment for the Deployment of Autonomous Shuttles},'' vol. 2675, no.~11, pp. 131--140, 2021. [Online]. Available: \url{www.pegasusprojekt.de}
\BIBentrySTDinterwordspacing

\bibitem{Maehara2021CALIBRATIONSHUTTLE}
S.~Maehara, ``{Calibration of Car-Following Models in the Presence of Autonomous Transit Shuttle},'' Ph.D. dissertation, University of Florida, Gainesville, 2021.

\bibitem{Treiber2000CongestedSimulations}
\BIBentryALTinterwordspacing
M.~Treiber, A.~Hennecke, and D.~Helbing, ``{Congested traffic states in empirical observations and microscopic simulations},'' \emph{Physical Review E}, vol.~62, no.~2, p. 1805, 8 2000. [Online]. Available: \url{https://journals.aps.org/pre/abstract/10.1103/PhysRevE.62.1805}
\BIBentrySTDinterwordspacing

\bibitem{Kesting2008CalibratingData}
\BIBentryALTinterwordspacing
A.~Kesting and M.~Treiber, ``{Calibrating Car-Following Models by Using Trajectory Data},'' \emph{Transportation Research Record: Journal of the Transportation Research Board}, vol. 2088, no.~1, pp. 148--156, 1 2008. [Online]. Available: \url{http://journals.sagepub.com/doi/10.3141/2088-16}
\BIBentrySTDinterwordspacing

\bibitem{Kesting2010EnhancedCapacity}
\BIBentryALTinterwordspacing
A.~Kesting, M.~Treiber, and D.~Helbing, ``{Enhanced intelligent driver model to access the impact of driving strategies on traffic capacity},'' \emph{Philosophical Transactions of the Royal Society A}, vol. 368, no. 1928, pp. 4585--4605, 10 2010. [Online]. Available: \url{https://royalsocietypublishing.org/doi/10.1098/rsta.2010.0084}
\BIBentrySTDinterwordspacing

\bibitem{Bu2010DesignSystem}
F.~Bu, H.~S. Tan, and J.~Huang, ``{Design and field testing of a Cooperative Adaptive Cruise Control system},'' \emph{Proceedings of the 2010 American Control Conference, ACC 2010}, pp. 4616--4621, 2010.

\bibitem{Milanes2014CooperativeSituations}
V.~Milanes, S.~E. Shladover, J.~Spring, C.~Nowakowski, H.~Kawazoe, and M.~Nakamura, ``{Cooperative adaptive cruise control in real traffic situations},'' \emph{Transactions on Intelligent Transportation Systems}, vol.~15, pp. 296--305, 2014.

\bibitem{Milanes2014ModelingData}
\BIBentryALTinterwordspacing
V.~Milan{\'{e}}s and S.~E. Shladover, ``{Modeling cooperative and autonomous adaptive cruise control dynamic responses using experimental data},'' \emph{Transportation Research Part C: Emerging Technologies}, vol.~48, pp. 285--300, 11 2014. [Online]. Available: \url{https://linkinghub.elsevier.com/retrieve/pii/S0968090X14002447}
\BIBentrySTDinterwordspacing

\bibitem{Melson2018DynamicControl}
C.~L. Melson, M.~W. Levin, B.~E. Hammit, and S.~D. Boyles, ``{Dynamic traffic assignment of cooperative adaptive cruise control},'' \emph{Transportation Research Part C: Emerging Technologies}, vol.~90, pp. 114--133, 2018.

\bibitem{Vanderwerf2002EffectsCapacity}
J.~Vanderwerf, S.~E. Shladover, M.~A. Miller, and N.~Kourjanskaia, ``{Effects of adaptive cruise control systems on highway traffic flow capacity},'' \emph{TRR: Journal of the TRB}, vol. 1800, no.~1, p. 78 – 84, 2002.

\bibitem{Jiang2001FullTheory}
\BIBentryALTinterwordspacing
R.~Jiang, Q.~Wu, and Z.~Zhu, ``{Full velocity difference model for a car-following theory},'' \emph{Physical Review E - Statistical Physics, Plasmas, Fluids, and Related Interdisciplinary Topics}, vol.~64, no.~1, p.~4, 2001. [Online]. Available: \url{https://www.researchgate.net/publication/11881499_Full_velocity_difference_model_for_car-following_theory}
\BIBentrySTDinterwordspacing

\bibitem{Kim2023IdentifyingHighways}
B.~Kim and K.~P. Heaslip, ``{Identifying suitable car-following models to simulate automated vehicles on highways},'' \emph{International Journal of Transportation Science and Technology}, vol.~12, no.~2, pp. 652--664, 6 2023.

\bibitem{Tiernan2019CARMAApplications}
T.~Tiernan, P.~Bujanovic, P.~Azeredo, W.~G. Najm, and T.~Lochrane, ``{CARMA Testing and Evaluation of Research Mobility Applications},'' Tech. Rep., 2019.

\bibitem{Fremont2020FormalWorld}
D.~J. Fremont, E.~Kim, Y.~V. Pant, S.~A. Seshia, A.~Acharya, X.~Bruso, P.~Wells, S.~Lemke, Q.~Lu, and S.~Mehta, ``{Formal Scenario-Based Testing of Autonomous Vehicles: From Simulation to the Real World},'' in \emph{2020 IEEE 23rd International Conference on Intelligent Transportation Systems, ITSC 2020}.\hskip 1em plus 0.5em minus 0.4em\relax Institute of Electrical and Electronics Engineers Inc., 9 2020.

\bibitem{Treiber2013}
M.~Treiber and A.~Kesting, ``{Microscopic Calibration and Validation of Car-Following Models – A Systematic Approach},'' \emph{Procedia - Social and Behavioral Sciences}, vol.~80, pp. 922--939, 6 2013.

\bibitem{deWinkel2023StandardsJerk}
K.~N. de~Winkel, T.~Irmak, R.~Happee, and B.~Shyrokau, ``{Standards for passenger comfort in automated vehicles: Acceleration and jerk},'' \emph{Applied Ergonomics}, vol. 106, p. 103881, 1 2023.

\bibitem{Kummetha2023ExaminingTrajectories}
V.~C. Kummetha, A.~Khoda~Bakhshi, S.~Concas, A.~Kourtellis, and A.~Mohammadnazar, ``{Examining Individualistic Driving Volatility Changes During and One Year into the COVID-19 Pandemic Using Paneled Connected Vehicle Trajectories},'' \emph{Transportation Research Record: Journal of the Transportation Research Board}, p. 036119812311583, 3 2023.

\bibitem{Punzo2012CanTrusted}
\BIBentryALTinterwordspacing
V.~Punzo, B.~Ciuffo, and M.~Montanino, ``{Can Results of car-following Model Calibration Based on Trajectory Data be Trusted?}'' \emph{https://doi.org/10.3141/2315-02}, vol. 2315, no. 2315, pp. 11--24, 1 2012. [Online]. Available: \url{https://journals.sagepub.com/doi/10.3141/2315-02}
\BIBentrySTDinterwordspacing

\bibitem{Punzo2021AboutCodes}
V.~Punzo, Z.~Zheng, and M.~Montanino, ``{About calibration of car-following dynamics of automated and human-driven vehicles: Methodology, guidelines and codes},'' \emph{Transportation Research Part C: Emerging Technologies}, vol. 128, 7 2021.

\bibitem{Hammit2018AData}
\BIBentryALTinterwordspacing
B.~Hammit, R.~James, and M.~Ahmed, ``{A Case for Online Traffic Simulation: Systematic Procedure to Calibrate Car-Following Models Using Vehicle Data},'' in \emph{2018 21st International Conference on Intelligent Transportation Systems (ITSC)}.\hskip 1em plus 0.5em minus 0.4em\relax IEEE, 11 2018, pp. 3785--3790. [Online]. Available: \url{https://ieeexplore.ieee.org/document/8569684/}
\BIBentrySTDinterwordspacing

\bibitem{Bae2019TowardBus}
I.~Bae, J.~Moon, and J.~Seo, ``{Toward a comfortable driving experience for a self-driving shuttle bus},'' \emph{Electronics (Switzerland)}, vol.~8, no.~9, 9 2019.

\bibitem{Kesting2008AdaptiveAvoidance}
\BIBentryALTinterwordspacing
A.~Kesting, M.~Treiber, M.~Sch{\"{o}}nhof, and D.~Helbing, ``{Adaptive cruise control design for active congestion avoidance},'' \emph{Transportation Research Part C: Emerging Technologies}, vol.~16, no.~6, pp. 668--683, 12 2008. [Online]. Available: \url{https://linkinghub.elsevier.com/retrieve/pii/S0968090X08000028}
\BIBentrySTDinterwordspacing

\bibitem{Zhu2018ModelingStudy}
M.~Zhu, X.~Wang, A.~Tarko, and S.~Fang, ``{Modeling car-following behavior on urban expressways in Shanghai: A naturalistic driving study},'' \emph{Transportation Research Part C: Emerging Technologies}, vol.~93, pp. 425--445, 8 2018.

\bibitem{Kovacs2016ParametriKrizanjima}
T.~Kov{\'{a}}cs, K.~Bolla, R.~A. Gil, E.~Csizm{\'{a}}s, C.~F{\'{a}}bi{\'{a}}n, L.~Kov{\'{a}}cs, K.~Medgyes, J.~Oszt{\'{e}}nyi, and A.~V{\'{e}}gh, ``{Parametri modela inteligentnog voza{\v{c}}a u signaliziranim kri{\v{z}}anjima},'' \emph{Tehnicki Vjesnik}, vol.~23, no.~5, pp. 1469--1474, 10 2016.

\bibitem{Salles2022ExtendingMeasurements}
D.~Salles, S.~Kaufmann, and H.-C. Reuss, ``{Extending the Intelligent Driver Model in SUMO and Verifying the Drive Off Trajectories with Aerial Measurements},'' \emph{SUMO Conference Proceedings}, vol.~1, pp. 1--25, 6 2022.

\bibitem{Xiao2017RealisticVehicles}
L.~Xiao, M.~Wang, and B.~van Arem, ``{Realistic car-following models for microscopic simulation of adaptive and cooperative adaptive cruise control vehicles},'' \emph{TRR: Journal of the TRB}, vol. 2623, pp. 1--9, 2017.

\end{thebibliography}

\bibliographystyle{IEEEtran}

\end{document}